\documentclass[10pt,twocolumn,letterpaper]{article}

\usepackage[pagenumbers]{cvpr} 

\def\confName{3DV\xspace}
\def\confYear{2024\xspace}
\def\paperID{47\xspace}

\newcommand{\methodname}{\methodCOLOR{TeCH}\xspace}
\newcommand{\mn}{\methodname}

\newcommand{\mnvqa}{\methodCOLOR{$\text{TeCH}_\text{vqa}$}\xspace}
\newcommand{\mndb}{\methodCOLOR{$\text{TeCH}_\text{db}$}\xspace}

\newcommand{\ourTitle}{Text-guided Reconstruction of Lifelike Clothed Humans}

\newcommand{\website}{\mbox{\href{https://huangyangyi.github.io/TeCH}{\tt\textit{huangyangyi.github.io/TeCH}}}}
\newcommand{\video}{\href{https://youtu.be/SjzQ6158Pho}{video}\xspace}

\usepackage{xcolor}
\usepackage{colortbl}
\usepackage{url}

\usepackage[pagebackref,breaklinks,colorlinks,citecolor=citecolor]{hyperref}
\usepackage{amsmath}
\usepackage{multirow}
\usepackage{float}
\usepackage{cuted}
\usepackage{bm}
\usepackage{comment}
\usepackage{caption}
\usepackage{balance}
\usepackage{pifont}
\usepackage[normalem]{ulem}
\usepackage{arydshln}
\usepackage{lipsum}
\usepackage{pifont}
\usepackage{stackengine}
\usepackage{fontawesome}
\usepackage[toc,page]{appendix}
\usepackage[accsupp]{axessibility}  

\definecolor{bittersweet}{rgb}{1.0, 0.44, 0.37}
\newcommand{\methodCOLOR}[1]{{\color{black} #1}}
\newcommand{\supmatCOLOR}[1]{{\color{red} #1}}

\definecolor{bestcolor}{rgb}{1, 0.5, 0.25}
\definecolor{secondbestcolor}{rgb}{1, 0.8, 0.5}
\newcommand{\bone}{\cellcolor{bestcolor}}
\newcommand{\btwo}{\cellcolor{secondbestcolor}}

\newcommand{\supmat}{\mbox{\supmatCOLOR{Appx.}}\xspace}

\definecolor{todocolor}{RGB}{255,0,00}

\definecolor{DeltaColor}{rgb}{0.039,0.73,0.71}
\definecolor{SigmaColor}{rgb}{0.98,0.45,0.0}
\definecolor{TODOColor}{rgb}{0.98,0.0,0.0}
\definecolor{AlphaColor}{rgb}{0,0,0.8}
\definecolor{BetaColor}{rgb}{0.8,0,0.8}
\definecolor{GammaColor}{rgb}{0.514,0.34,0.224}
\definecolor{EpsilonColor}{rgb}{0.353,0.725,0.906}
\definecolor{citecolor}{HTML}{0071bc}

\newcommand{\yangyi}[1]{{\color{BetaColor} Yangyi: #1 }}

\let\footnotesize=\scriptsize  
\newcommand{\qheading}[1]{\noindent\textbf{#1}.}
\newcommand{\mheading}[1]{\medskip\noindent\textbf{#1}.}
\newcommand{\sheading}[1]{\smallskip\noindent\textbf{#1}.}
\newcommand{\zheading}[1]{\textbf{#1}.}

\newlength\savewidth\newcommand\shline{\noalign{\global\savewidth\arrayrulewidth
  \global\arrayrulewidth 1pt}\hline\noalign{\global\arrayrulewidth\savewidth}}

\DeclareTextFontCommand{\specific}{\small\fontfamily{qcr}\selectfont}

\newcommand{\colorRef}[1]{\textcolor{red}{#1}} 
\usepackage[capitalize]{cleveref}
\crefname{figure}{\colorRef{Fig.}}{\colorRef{Figs.}}
\Crefname{figure}{\colorRef{Figure}}{\colorRef{Figures}}
\crefname{section}{\colorRef{Sec.}}{\colorRef{Secs.}}
\Crefname{section}{\colorRef{Section}}{\colorRef{Sections}}
\Crefname{table}{\colorRef{Table}}{\colorRef{Tables}}
\crefname{table}{\colorRef{Tab.}}{\colorRef{Tabs.}}

\newcommand{\ourrefcolor}[1]{\textcolor{red}{#1}} 

\newcommand{\Tab}[1]{\mbox{\ourrefcolor{{Table}}~\ref{#1}}}

\newcommand{\eref}[1]{Eq.~(\ref{#1})}

\newcommand{\sref}[1]{Sec.~\ref{#1}}
\author{
    Yangyi Huang\textsuperscript{1}$^{*}$,
    Hongwei Yi\textsuperscript{2}$^{*}$,
    Yuliang Xiu\textsuperscript{2}$^{*}$,
    Tingting Liao\textsuperscript{3},
    Jiaxiang Tang\textsuperscript{4},
    Deng Cai\textsuperscript{1},
    Justus Thies\textsuperscript{2}
    \\
    {\normalsize\textsuperscript{1}State Key Lab of CAD \& CG, Zhejiang University
    \quad
    \textsuperscript{2}Max Planck Institute for Intelligent Systems}
    \\
    {\normalsize\textsuperscript{3}Mohamed bin Zayed University of Artificial Intelligence
    \quad
    \textsuperscript{4}Peking University}
    \\
    {\tt\small huangyangyi@zju.edu.cn, \{hongwei.yi, yuliang.xiu, justus.thies\}@tuebingen.mpg.de}\\ 
    {\tt\small tingting.liao@mbzuai.ac.ae, tjx@pku.edu.cn, dengcai@cad.zju.edu.cn}\\
}

\newcommand{\smplx}{\mbox{SMPL-X}\xspace}

\newcommand{\smpl}{\mbox{SMPL}\xspace}

\newcommand{\thtwo}{\mbox{THuman2.0}\xspace}
\newcommand{\cape}{\mbox{CAPE}\xspace}
\newcommand{\laion}{\mbox{LAION-5B}\xspace}

\newcommand{\norm}[1]{\left\lVert#1\right\rVert}

\newcommand{\cmark}{\ding{51}}%
\newcommand{\xmark}{\color{red}{\ding{55}}}%

\newcommand{\pixie}{\mbox{PIXIE}\xspace}

\newcommand{\db}{\mbox{DreamBooth}\xspace}
\newcommand{\df}{\mbox{DreamFusion}\xspace}
\newcommand{\dmtet}{\mbox{DMTet}\xspace}

\newcommand{\sd}{\mbox{Stable Diffusion}\xspace}

\newcommand{\segf}{\mbox{SegFormer}\xspace}

\newcommand{\fd}{\mbox{Fantasia3D}\xspace}
\newcommand{\pifu}{\mbox{PIFu}\xspace}
\newcommand{\pamir}{\mbox{PaMIR}\xspace}
\newcommand{\phorhum}{\mbox{PHORHUM}\xspace}
\newcommand{\pifuhd}{\mbox{PIFuHD}\xspace}
\newcommand{\econ}{\mbox{ECON}\xspace}
\newcommand{\icon}{\mbox{ICON}\xspace}

\DeclareSymbolFont{matha}{OML}{txmi}{m}{it}
\DeclareMathSymbol{\varv}{\mathord}{matha}{118}

\begin{document}

\title{\mn: \ourTitle}

\newcommand{\teaserCaption}{
Given a single image, \mn reconstructs a lifelike 3D clothed human. 
\textbf{``Lifelike''} refers to 1) a detailed full-body geometry, including facial features and clothing wrinkles, in both frontal and unseen regions, and 2) a high-quality texture with consistent color and intricate patterns. 
The key insight is to guide the reconstruction using a personalized Text-to-Image (T2I) diffusion model and textual information derived via visual questioning answering (VQA).
Multi-view supervision is established through Score Distillation Sampling (SDS).
%
}

\twocolumn[{
    \renewcommand\twocolumn[1][]{#1}
    \maketitle
    \centering
    \begin{minipage}{1.00\textwidth}
        \centering 
        \includegraphics[trim=000mm 12mm 000mm 000mm, clip=True, width=\linewidth]{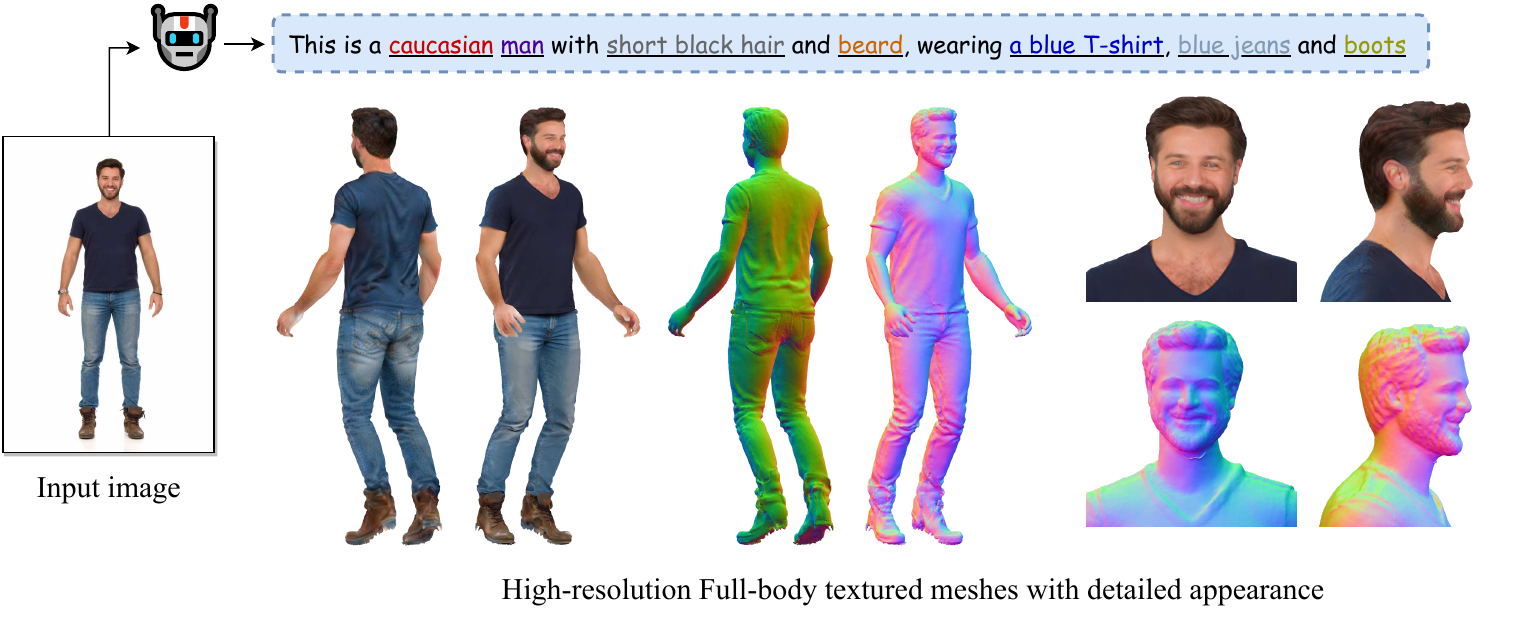}
    \end{minipage}
    \captionsetup{type=figure}
    \captionof{figure}{\looseness=-1{\teaserCaption}}
    \label{fig:teaser}
    \vspace{2.0 em}
}]

\def\thefootnote{*}\footnotetext{These authors contributed equally to this work.}
\begin{abstract}
Despite recent research advancements in reconstructing clothed humans from a single image, accurately restoring 
 the ``unseen regions'' with high-level details remains an unsolved challenge that lacks attention. Existing methods often generate overly smooth back-side surfaces with a blurry texture.
%
But how to effectively capture all visual attributes of an individual from a single image, which are sufficient to reconstruct unseen areas (\eg the back view)?
Motivated by the power of foundation models, \mn reconstructs the 3D human by leveraging 1) descriptive text prompts (\eg garments, colors, hairstyles) which are automatically generated via a garment parsing model and Visual Question Answering (VQA), 2) a personalized fine-tuned Text-to-Image diffusion model (T2I) which learns the ``indescribable'' appearance.
%
%
To represent high-resolution 3D clothed humans at an affordable cost, we propose a hybrid 3D representation based on \dmtet, which consists of an explicit body shape grid and an implicit distance field.
Guided by the descriptive prompts + personalized T2I diffusion model, the geometry and texture of the 3D humans are optimized through multi-view Score Distillation Sampling (SDS) and reconstruction losses based on the original observation.
%
%
%
\mn produces high-fidelity 3D clothed humans with consistent \& delicate texture, and detailed full-body geometry. 
%
Quantitative and qualitative experiments demonstrate that \mn outperforms the state-of-the-art methods in terms of reconstruction accuracy and rendering quality.
%
The code will be publicly available for research purposes at \website
\end{abstract}
\section{Introduction} 
\label{sec:introduction}

High-fidelity 3D digital humans are crucial for various applications in augmented and virtual reality, such as gaming, 
social media, 
education, 
e-commerce, 
and immersive telepresence. 
To facilitate the creation of digital humans from easily accessible in-the-wild photos, numerous approaches focus on reconstructing a 3D clothed human shape from a single image~\cite{saito2019pifu, saito2020pifuhd, he2020geoPifu, xiu2022icon, xiu2022econ, li2020monoport, li2020monoportRTL, huangARCHAnimatableReconstruction2020, heARCHAnimationReadyClothed2021, cao2022jiff,zheng2021pamir, saharia2022photorealistic,liao2023car}.
However, despite the advancements made by previous approaches, this specific problem can be considered ill-posed due to the lack of observations of non-visible areas.
Efforts to predict \textit{invisible} regions (\eg back-side) based on \textit{visible} visual cues (\eg colors~\cite{saito2019pifu,huangARCHAnimatableReconstruction2020,alldieck2022photorealistic}, normal estimates~\cite{xiu2022icon,xiu2022econ,saito2020pifuhd}) have proven unsuccessful, resulting in blurry texture and smoothed-out geometry, see~\cref{fig:qualitative-wild}.
%
%
As a result, inconsistencies arise when observing these reconstructions from different angles.
To address this issue, introducing multi-view supervision could be a potential solution. But is it feasible given only a single input image?
Here, we propose \mn to answer this question. Unlike prior research that primarily explores the connection between visible frontal cues and non-visible regions, \mn integrates textual information derived from the input image with a personalized Text-to-Image diffusion model, \ie, \db~\cite{ruiz2022dreambooth}, to guide the reconstruction process.
%
%
%
%
%

%
Specifically, we divide the information from the single input image into the semantic information that can be accurately described by texts and subject's distinctive and fine-detailed appearance which is not easily describable by text: 

\noindent \textbf{1) Describable} semantic prompts, including the detailed descriptions of colors, styles of garments, hairstyles, and facial features, are \textit{explicitly} parsed from the input image using a garment parsing model (\ie~\segf~\cite{xie2021segformer}) and a pre-trained visual-language VQA model (\ie BLIP~\cite{li2022blip}).
%
%

\noindent \textbf{2) Indescribable} appearance information, which \textit{implicitly} specifies the subject's distinctive appearance and fine-grained details, is embedded into a unique token ``[$V$]'', by a personalized Text-to-Image (T2I) diffusion model~\cite{ruiz2022dreambooth}.
%
%

%
Based on these information sources, we optimize the 3D human using multi-view Score Distillation Sampling (SDS)\cite{poole2022dreamfusion}, reconstruction losses based on the original observations, and regularization obtained from off-the-shelf normal estimators, to enhance the fidelity of the reconstructed 3D human models while preserving their original identity.
To represent a high resolution geometry at an affordable cost, we propose a hybrid 3D representation based on \dmtet~\cite{shen2021deep, gao2020learning}.
This hybrid 3D representation combines an explicit tetrahedral grid to approximate the overall body shape and implicit Signed Distance Function (SDF) and RGB fields to capture fine details in geometry and texture.
%
In a two-stage optimization process, we first optimize this tetrahedral grid, extract the geometry represented as a mesh, and then optimize the texture.
%

%
%
%
\mn enables the reconstruction of high-fidelity 3D clothed humans with detailed full-body geometry, and intricate textures with consistent color and patterns. As a result, it facilitates various downstream applications such as novel view rendering, character animation, and shape \& texture editing.
%
%
%
Quantitative evaluations performed on 3D clothed human datasets, covering various poses (CAPE~\cite{Pons-Moll:Siggraph2017}) and outfits (\thtwo~\cite{tao2021function4d}), have demonstrated \mn's superiority in reconstructing geometric details.
Qualitative comparisons conducted on in-the-wild images, accompanied by a perceptual study, further confirm that \mn surpasses SOTA methods in terms of rendering quality.
The code will be publicly avaiable for research purpose at \website

\newcommand{\methodCaption}{
\textbf{Method overview.} \mn takes an image $\mathcal{I}$ of a human as input. 
Text guidance is constructed through \textbf{(a)} using garment parsing model (\segf) and VQA model (BLIP) to parse the human attributes $A$ with pre-defined problems $Q$, and \textbf{(b)} embedding with subject-specific appearance into \db~$\mathcal{D'}$ as unique token $[V]$. 
Next, \mn represents the 3D clothed human with \textbf{(c)} \smplx initialized hybrid \dmtet, and optimize both geometry and texture using $\mathcal{L}_\text{SDS}$ guided by prompt $P=[V]+P_\text{VQA}(A)$. 
During the optimization, $\mathcal{L}_\text{recon}$ is introduced to ensure input view consistency, $\mathcal{L}_\text{CD}$ is to enforce the color consistency between different views, and $\mathcal{L}_\text{normal}$ serves as surface regularizer. Finally, the extracted  high-quality textured meshes \textbf{(d)} are ready to be used in various downstream applications.
}

\begin{figure*}[tbp]
    \centering
    \scriptsize
    \includegraphics[width=\linewidth]{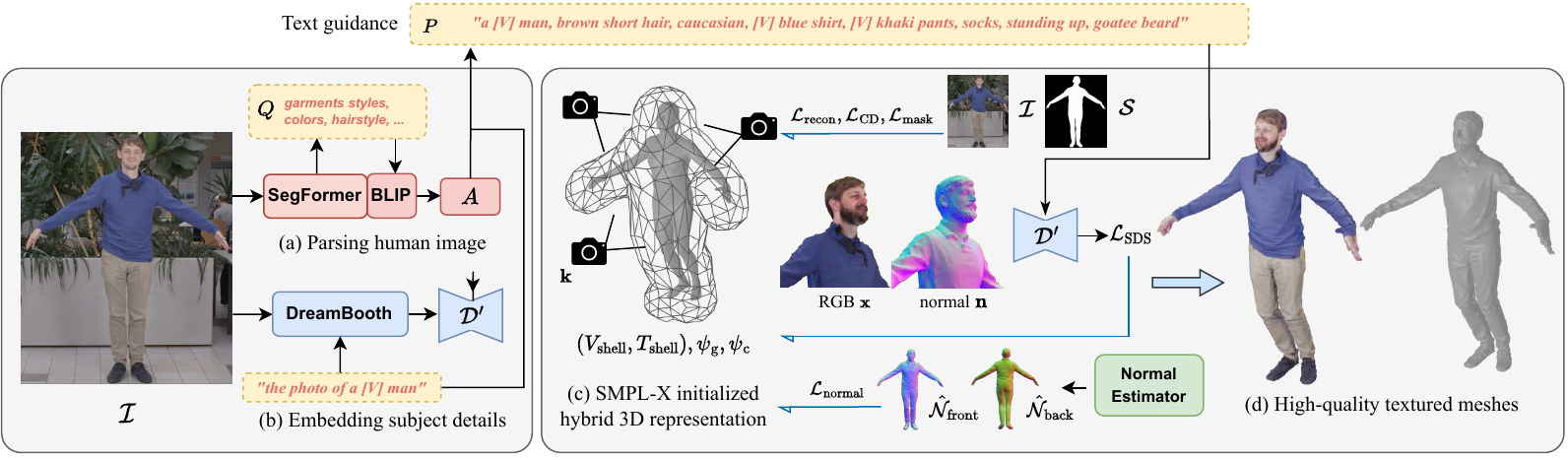}
    \captionof{figure}{\methodCaption}
    \label{fig:method}
\end{figure*}

\section{Related Work} \label{sec:relatedwork}

\mn \textbf{reconstructs} a high-fidelity clothed human from a single image, and \textbf{imagine} the missing parts through the aid of descriptive prompts and a personalized diffusion model. 
We relate \mn to both image-based human reconstructors (\cref{sec:related-reconstructor}) and 3D human generators (\cref{sec:related-generator}). 
Human reconstructors could be grouped as: 1) Explicit-shape-based, 2) Implicit-function-based, and 3) NeRF-based methods. 
The human generators are categorized \wrt their training data: 1) directly learned from 3D real captures or 2) indirectly learned from large-scale 2D images.  
In addition, there is a line of image-to-3D works focusing on general objects, which will be discussed in \cref{sec:img-to-3d}. 

\subsection{Image-based Clothed Human Reconstruction}
\label{sec:related-reconstructor}

\mheading{Explicit-shape-based Methods}
Human Mesh Recovery (HMR) from a single RGB image is a long-standing problem that has been thoroughly explored.
A lot of methods~\cite{pymaf2021, feng2021pixie, li2022cliff, li2021hybrik, Kanazawa2018_hmr, pare,spec,pymafx2023,li2023niki,VIBE:CVPR:2020,Kolotouros2019_spin} use mesh-based parametric body models~\cite{SMPL:2015, xu2020ghum, pavlakos2019expressive,Joo_2018_CVPR} to regress the shape and pose of minimally-clothed 3d body meshes.
To account for the 3D garments, 3D clothing offsets~\cite{alldieckTex2ShapeDetailedFull2019, alldieck2018dv,alldieck2018videoavatar,alldieck2019peopleInClothing,zhu2019hierarchMeshDeform,lazova2019textures360,xiang2020monoClothCap} 
or deformable garment templates~\cite{bhatnagar2019multiGarmentNet,jiang2020bcnet} are used on top of a body model.
Also, non-parametric explicit representations, such as depth maps~\cite{smith2019facsimile,gabeur2019moulding}, normal maps~\cite{xiu2022econ}, and point clouds~\cite{zakharkin2021point} could be leveraged to reconstruct the clothed human.
However, explicit shapes often suffer from restricted topological flexibility, particularly, when dealing with outfit variations in real-world scenarios, \eg, dress, skirt, and open jackets.
%
%

\mheading{Implicit-function-based Methods}
%
Implicit representations (occupancy/distance field) are topology-agnostic, thus, can represent 3D clothed humans, with arbitrary topologies, such as open jackets and loose skirts.
A line of works regresses the free-form implicit surface in an end-to-end manner~\cite{saito2019pifu,saito2020pifuhd,alldieck2022photorealistic}, leverages a 3D geometric prior~\cite{he2020geoPifu,corona2023s3f,xiu2022icon,zheng2021pamir,huangARCHAnimatableReconstruction2020,heARCHAnimationReadyClothed2021,cao2022jiff,liao2023car,yang2023dif}, or progressively builds up the 3D human using a ``sandwich-like'' structure and implicit shape completion~\cite{xiu2022econ}.
Among these works, \pifu~\cite{saito2019pifu}, ARCH(++)~\cite{huangARCHAnimatableReconstruction2020,heARCHAnimationReadyClothed2021}, and \pamir~\cite{zheng2021pamir} infer the full texture from the input image. 
\phorhum~\cite{alldieck2022photorealistic} and S3F~\cite{corona2023s3f} additionally decompose the albedo and global illumination.
%
%
However, the lack of multi-view supervision often results in depth ambiguities or inconsistent textures.
%
%
%

\mheading{NeRF-based Methods}
There is a separate line of research that focuses on optimizing neural radiance fields (NeRF) from a single image.
%
%
SHERF~\cite{hu2023sherf} and ELICIT~\cite{huang2022one} optimize a generalized human NeRF, incorporating model-based priors (SMPL-X). 
While SHERF complements missing information from partial 2D observations, ELICIT utilizes pre-trained CLIP~\cite{clip} to provide an appearance prior.
%
%

\subsection{Generative Modeling of 3D Clothed Humans}
\label{sec:related-generator}

\zheading{3D Human Generator Trained on 3D Data}
Statistical body models~\cite{SMPL:2015, xu2020ghum, pavlakos2019expressive,Joo_2018_CVPR} can be considered as 3D generative models of the human body.
These models are trained on numerous 3D scans of minimally-clothed bodies, and can generate posed bodies with varying shapes, but without clothing. 
To account for the outfits, CAPE~\cite{CAPE:CVPR:20} learns a clothing offset layer based on the SMPL-D model, from registered human scans, Chupa~\cite{kim2023chupa} ``carves'' the \smpl mesh by dual normal maps generated by pose-conditioned diffusion model; Alternatively, gDNA~\cite{chen2022gdna}, NPMs~\cite{palafox2021npms}, and SPAMs~\cite{palafox2021spams}, learn the implicit clothed avatars from normalized raw captures (\ie, scans, depth maps).
%
%
Unfortunately, all the aforementioned methods to learn generative 3D humans with diverse shapes and appearances require 3D data, which is both limited and expensive to acquire. 
Rodin~\cite{wang2022rodin} has recently employed large-scale 3D synthetic head avatars in combination with a diffusion model to develop a high-fidelity head avatar generator.
However, the scarcity of datasets containing real 3D clothed humans~\cite{Zheng2019DeepHuman, tao2021function4d, 2023dnarendering, isik2023humanrf,cai2022humman} limits the model's generalization ability and may lead to overfitting on small datasets.

\mheading{3D Human Generator from 2D Image Collections}
In contrast to 3D data, large-scale 2D human images are widely avaible from DeepFashion~\cite{liuLQWTcvpr16DeepFashion,DeepFashion2}, SHHQ~\cite{fu2022styleganhuman} and \laion~\cite{schuhmann2022laionb}.
%
%
Related human generators represent 3D humans using meshes~\cite{grigorev2021stylepeople,hong2022avatarclip,jiang2023avatarcraft}, \dmtet~\cite{gao2022get3d}, Tri-planes~\cite{noguchi2022unsupervised, zhang2023avatargen, bergman2022generative,jiang2022humangen,dong2023ag3d}, implicit functions~\cite{xiong2023get3dhuman}, or neural fields~\cite{hongEVA3DCompositional3D2022,cao2023dreamavatar,kolotouros2023dreamhuman,zeng2023avatarbooth}. 
%
%
Some methods adapt GANs~\cite{Karras2019stylegan2} by integrating diff-renderer~\cite{grigorev2021stylepeople,noguchi2022unsupervised, zhang2023avatargen, bergman2022generative,jiang2022humangen,xiong2023get3dhuman,dong2023ag3d,svitov2023dinar}, while others leverage diffusion models~\cite{cao2023dreamavatar,hong2022avatarclip,kolotouros2023dreamhuman,huang2023dreamwaltz,zhang2023avatarverse}. 
Despite the demonstrated quality of these methods in generating textured avatars, a significant gap still exists in achieving ``lifelike'' avatars with detailed geometry and texture, consistent with the input.
%

%
In contrast, \mn excels at generating ``lifelike'' 3D characters from a single image, incorporating consistent texture with intricate patterns like checkered or overlapped designs.
It relies on a pretrained diffusion model which is trained on a billion-level data, \laion~\cite{schuhmann2022laionb}, and offers the ability to \textbf{imagine the non-visible regions}, guided by descriptive prompts.
%
%
Furthermore, it leverages the image-based reconstruction approach to faithfully \textbf{reconstruct the visible regions} from a single input image.

\subsection{Image-to-3D for General Objects}
\label{sec:img-to-3d}

Lifting 2D to 3D for general objects is a longstanding problem with valuable explorations. Here, we mainly focus on diffusion-guided approaches. Initially, CLIP~\cite{clip} semantic consistency loss~\cite{Jain2021DietNeRF}, Score Jacobian Chaining (SJC)~\cite{wang2022score} and Score Distillation Sampling (SDS)~\cite{poole2022dreamfusion} are proposed to leverage pretrained 2D diffusion models for 3D content generation. 
Subsequently, there is a line of works~\cite{raj2023dreambooth3d,deng2022nerdi,xu2022neurallift,tang2023make, melaskyriazi2023realfusion} that address this problem, by incorporating textural inversion \cite{gal2022image}, \db~\cite{ruiz2022dreambooth}, CLIP-guided diffusion prior, depth prior, and reconstruction loss. 
In addition to aforementioned ``reconstruct via multi-view SDS'' scheme, recent attention has been drawn to the ``reconstruct via direct view-conditional generation''~\cite{watson2023threedim,qian2023magic123,zhou2023sparsefusion,liu2023one,liu2023zero1to3, chan2023genvs}. 
In contrast, \mn aims to recover pixel-aligned models with intricate texture, even in non-visible regions, which is a challenging scenario where existing solutions have not shown promising results.

 \section{Method} \label{sec:method} 
Given a single image as input, \mn aims at reconstructing a high-fidelity 3D clothed human.
Here, ``high-fidelity'' refers to the inclusion of consistent texture with intricate patterns, as well as detailed full-body geometry. 
To achieve this, \mn follows a two-step procedure:  
Firstly, a text prompt that describes the human in the input image is obtained via the human parsing model \segf~\cite{xie2021segformer} and the VQA model BLIP~\cite{li2022blip} (\sref{sec:text_guidance}).
This descriptive prompt is used to guide the generation process in \db~\cite{ruiz2022dreambooth}, a personalized Text-to-Image diffusion model fine-tuned on augmented input images.
Secondly, the 3D human, which is represented as hybrid \dmtet and initialized with \smplx (\sref{sec:representation}), is optimized with SDS losses~\cite{poole2022dreamfusion} computed from the personalized \db (\sref{sec:optimization}). 
The Score Distillation Sampling (SDS) loss has been introduced in \df~\cite{poole2022dreamfusion} for the task of Text-to-3D generation of general objects, by optimizing a neural radiance field (NeRF) with gradients from a frozen diffusion model. 
In our case, we utilize the SDS loss to guide the reconstruction of a 3D human from a single input image, employing a multi-stage optimization strategy (\sref{sec:optimization}) to get a consistent alignment of geometry and texture.

\subsection{Extracting Text-guidance from the Observation} 
\label{sec:text_guidance}

\begin{figure}[t]
\centering
\scriptsize
\includegraphics[width=\linewidth]{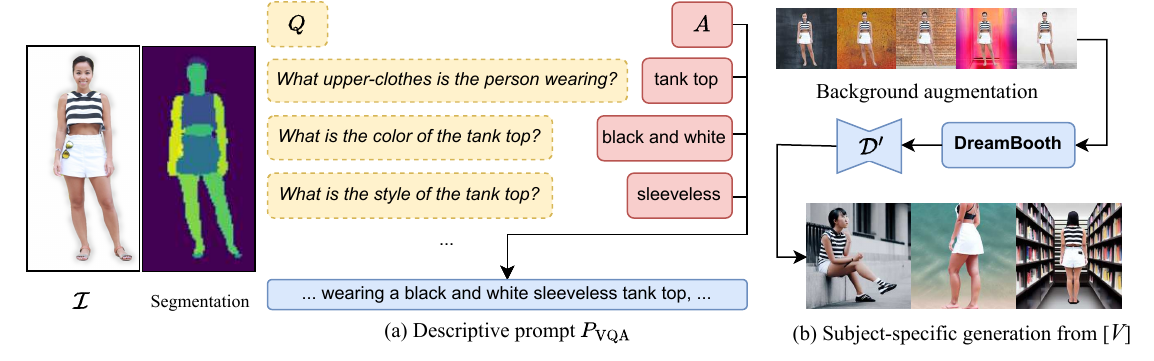}  
\scriptsize
\caption{\textbf{Prompt construction ($P = P_\mathrm{VQA} + [V]$).} 
(a) Inquire VQA model with predefined questions on individual appearance to construct \textit{describable} prompts $P_\mathrm{VQA}$. (b) Fine-tuned \db with background-augmented images to embed \textit{indescribable} subject-specific details into unique identifier $[V]$.
}
\label{fig:method-text-guidance}
\end{figure} 

\mheading{Parsing human attributes}
As depicted in ~\cref{fig:method-text-guidance}, given the input image of a human, \segf~\cite{xie2021segformer}, which is fine-tuned on ATR dataset~\cite{liang2015deep,liang2015human}, is applied to recognize each part of the garments (\eg hat, skirt, pants, belt, shoes).
%
To obtain detailed descriptions (\ie color and style) of the parsed garments, we utilize the vision-language model BLIP~\cite{li2022blip} as VQA captioner. This model has been pre-trained on a vast collection of image-text pairs, enabling it to automatically generate descriptive prompts.
Rather than using naive image captioning, we employ a series of fine-grained VQA questions $\{Q_i\}$ (see \supmat's~\cref{sec:vqa-questions}) as input to BLIP. These questions cover garment styles, colors, facial features, and hairstyles, with the corresponding answers denoted as $\{A_i\}$.
The set of $\{A_i\}$ will be inserted into a predefined template to create text prompts $P_\mathrm{VQA}$, which will serve as text-guidance to condition the text-to-image diffusion model, recap the full method overview in~\cref{fig:method}.

\begin{figure}[t]
\centering
\scriptsize
\includegraphics[width=\linewidth]{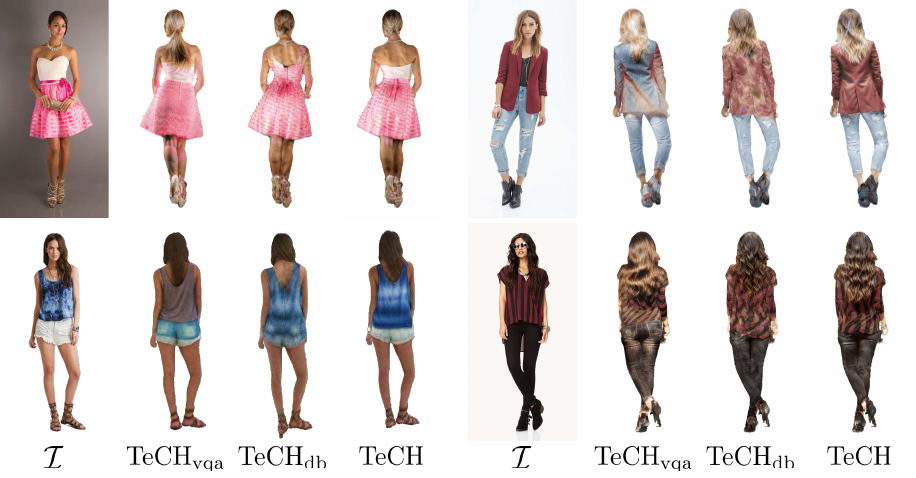}  
\scriptsize
\caption{\textbf{The effects of text guidance.} We compare the effectiveness of using only VQA descriptions (\mnvqa), only \db identity token (\mndb), and both of them (\mn).
}
\label{fig:ablation_semantic}
\end{figure}

\mheading{Embedding subject-specific appearance}
Does the text prompt $P_\mathrm{VQA}$ comprehensively capture all the visual characteristics of the subject? No, a picture is worth a thousand words.
Thus, we utilize \db~\cite{ruiz2022dreambooth} to learn the \textit{indescribable} visual appearance. 
DreamBooth is a method for ``personalizing'' a diffusion model through few-shot tuning (3$\sim$5 images).
We perform DreamBooth's fine-tuning on a pre-trained \sd(v1.5) as the base model.
%
To generate the needed inputs, we augment the single input image with five different backgrounds, as shown in~\cref{fig:method-text-guidance}. 
To prevent language drift, we assign the subject classes ``\specific{man}'' or ``\specific{woman}'' based on the gender determined by the VQA.
After fine-tuning \db, the subject-specific distinctive appearance is encoded within a unique identifier token ``$[V]$''.
We insert ``$[V]$'' into the prompt $P_\mathrm{VQA}$, to construct the final text prompt $P$ used by the personalized \db $\mathcal{D}'$. 
In~\cref{fig:ablation_semantic}, you can see how these individual prompts contribute to the final appearance. 
%

\begin{figure*}[ht]
\centering
\scriptsize
\includegraphics[width=\linewidth]{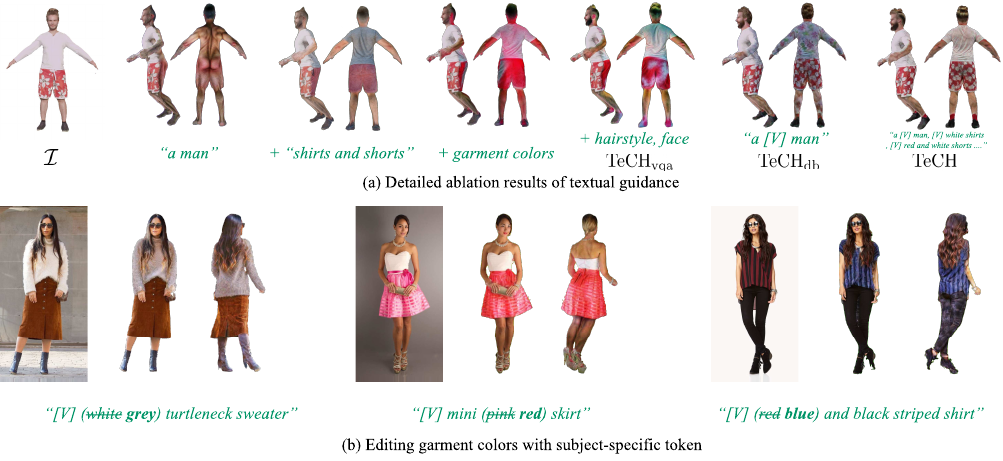}  
\scriptsize
\caption{\textbf{(a)} Top depicts the impact of specific elements within the textual guidance, such as garment styles \& colors, hairstyle, facial features, and the placement \& inclusion of ``$[V]$''. \textbf{(b)} Bottom demonstrates that \mn facilities text-guided garment color editing.
}
\label{fig:ablation_semantic_more}
\end{figure*} 

\mheading{Deeper analysis of description $P$}
In \cref{fig:ablation_semantic_more} (a), we first show the impact of individual elements within the text prompt, including garment styles \& colors, hairstyle, and face, which guide the model to recover the appearance of each attribute of the clothed human.
The first column shows that a basic class description alone cannot effectively guide the reconstruction process.
However, in the subsequent columns, text guidance incorporating detailed descriptions of clothing proves successful in accurately reconstructing the structure of clothed humans.
Furthermore, with additional information regarding colors and hairstyles, the characters reconstructed by \mnvqa exhibit greater semantic consistency with respect to the input view.
However, merely relying on VQA descriptions is insufficient for generating a ``convincingly fake'' appearance.
%

Only using the DreamBooth guidance (\mndb), helps to recover original garment patterns, which demonstrates that DreamBooth has a high-level understanding of texture patterns. 
However, it sometimes will \textit{diffuse} the patterns to the entire human.
By combining ``$[V]$'' with the VQA parsing text prompts $P_\mathrm{VQA}$, \mn produces remarkably realistic texture with consistent color and intricate patterns.
In \cref{fig:ablation_semantic_more} (b), we also demonstrate some text-guided garment color editing examples based on a fine-tuned \db model $D'$ and subject-specific token ``$[V]$''.
%
%


\subsection{Hybrid 3D Representation}
\label{sec:representation}
To efficiently represent the 3D clothed human at a high resolution, we embed \dmtet~\cite{gao2020learning, shen2021deep} around the \smplx body mesh~\cite{Onizuka2020}.
Specifically, we construct a compact tetrahedral grid $(V_\mathrm{shell}, T_\mathrm{shell})$ within an outer shell $M_\mathrm{shell}$, shown in~\cref{fig:method}-(c).
Compared to the \dmtet cubic-based tetrahedral grid, the outer shell tetrahedral grid is more computationally efficient for high-resolution geometry modeling of a human.
Using \pixie~\cite{feng2021pixie}, we estimate an initial body $\mathcal{M}_\mathrm{body}$.
To create $M_\mathrm{shell}$, a series of mesh dilation, down-sampling, and up-sampling steps are applied to the body mesh $M_\mathrm{body}$ (see details \cref{sec:outer-shell} of~\supmat).
We use two MLP networks $\Psi_\mathrm{g}, \Psi_\mathrm{c}$ with hash encoding~\cite{muller2022instant}, parameterized by $\psi_g$ and $\psi_c$ to learn the geometry and color separately.
The geometry network $\Psi_\mathrm{g}$ predicts the SDF value $\Psi_\mathrm{g}(v_i) = s(v_i; \psi_\mathrm{g})$ of each \dmtet vertex $v_i$.
It is initialized by fitting it to the SDF of $M_\mathrm{shell}$:
%
\begin{equation}
    \mathcal{L}_\mathrm{init}=\sum_{p_i\in \mathbf{P}_\mathrm{}}\norm{s(p_i; \psi_\mathrm{g}) - \mathrm{SDF}(p_i)}_2^2 ,
\end{equation}
where $\mathbf{P}=\{p_i\in\mathbb{R}^3\}$ is a point set randomly sampled near $M_\mathrm{shell}$, and $\mathrm{SDF}(p_i)$ is the pre-computed pointwise SDF.
Triangular meshes can be extracted from this efficient hybrid 3D representation by Marching Tetrahedra (MT)~\cite{doi1991efficient}:
\begin{equation}
M=\mathrm{MT}(V_\mathrm{shell}, T_\mathrm{shell}, s(V_\mathrm{shell};\psi_\mathrm{g})) .
\end{equation} 
Given the camera parameters $\mathbf{k}$, the generated mesh is rendered through differentiable rasterization $\mathcal{R}$~\cite{Laine2020diffrast}, to get the back-projected 3D locations $\mathcal{P}(M, \mathbf{k})$, rendered mask $\mathcal{M}(M, \mathbf{k})$, and rendered normal image $\mathcal{N}(M, \mathbf{k})$
\begin{equation}
    \mathcal{R}(M, \mathbf{k}) = \left(\mathcal{P}(M, \mathbf{k}), \mathcal{M}(M, \mathbf{k}), \mathcal{N}(M, \mathbf{k})\right)
\end{equation}
%
The albedo of each back-projected pixel is predicted by the color network $\Psi_\mathbf{c}$, where $\psi_\mathrm{c}$ represents the parameters:
\begin{equation}
    \mathcal{I'}(M, \psi_\mathrm{c}, \mathbf{k}) = \Psi_\mathbf{c}(\psi_\mathrm{c}, \mathcal{P}(M, \mathbf{k})) .
\end{equation}
%

As detailed in \Cref{sec:optimization}, we optimize this 3D representation using a coarse-to-fine strategy by applying successive subdivisions on the tetrahedral grids.
Specifically, a more detailed surface $M_\mathrm{subdiv}(\psi_\mathrm{g})$ can be obtained by applying volume subdivision on the surface tetrahedral grids $(V_\mathrm{surface}, T_\mathrm{surface})$ that intersect with $M(\psi_\mathrm{g})$. Note that the SDF values of the refined vertices are still inferred by $\Psi_\mathrm{g}$.
%




\subsection{Multi-stage Optimization}
\label{sec:optimization}

We adopt a multi-stage, coarse-to-fine optimization process to sequentially recover the subject's geometry and texture.
In the initial stage, we utilize the tetrahedral representation to model the subject's geometry (\cref{sec:geometry-stage}).
Next, the appearance is recovered using the mesh that is extracted from the tetrahedral grid (\cref{sec:texture-stage}).
Both stages are leveraging SDS-based losses using the personalized \db model which provides multi-view supervision
by sampling new camera views as described in \cref{sec:camera_sampling}.
%
%

\subsubsection{Geometry Stage}
\label{sec:geometry-stage}
We optimize the geometry based on a silhouette loss $\mathcal{L}_\mathrm{sil}$ using the orig. image, a text-guided SDS loss on rendered normal images $\mathcal{L}_\mathrm{SDS}^\mathrm{norm}$, and geometric regularization $\mathcal{L}_\mathrm{reg}$ based on pred. normals $\mathcal{L}_\mathrm{norm}$ and surface smoothness $\mathcal{L}_\mathrm{lap}$:
\begin{align}
\begin{split}
\label{overall_l}
    \mathcal{L}_\mathrm{geometry} &= \lambda_\mathrm{sil}\mathcal{L}_\mathrm{sil} + \lambda_\mathrm{SDS}\mathcal{L}_\mathrm{SDS}^\mathrm{norm} + \mathcal{L}_\mathrm{reg}\\ 
    \mathcal{L}_\mathrm{reg} & = \lambda_\mathrm{norm}\mathcal{L}_\mathrm{norm}  +\lambda_\mathrm{lap}\mathcal{L}_\mathrm{lap} , 
\end{split}
\end{align}
where $\lambda$ represents the weights to balance the losses.
During optimization of this loss, we perform a coarse-to-fine subdivision on \dmtet, to robustly produce a high-resolution mesh for the clothed body.
Specifically, the optimization is first performed w/o subdivision for $t_\mathrm{coarse}=5000$ iters, and then with subdivision for $t_\mathrm{fine}=5000$ iters. 

\mheading{Pixel-aligned silhouette loss} 
The silhouette loss~\cite{zhangPerceiving3DHumanObject2020, yiHumanAwareObjectPlacement2022} enforces pixel-alignment with the foreground mask $\mathcal{S}$ of the input image $\mathcal{I}$ under the input camera view $\mathbf{k}$:
\begin{align}
\begin{split}
    \mathcal{L}_\mathrm{sil} &= \norm{\mathcal{S} - \mathcal{M}(M, \mathbf{k})}_2^2 \\ &+ \sum_{x\in \text{Edge}(\mathcal{M}(M, \mathbf{k}))}\min_{\hat x\in \text{Edge}(\mathcal{S})}\norm{x-\hat{x}}_1 .
\end{split}
\end{align}
It consists of (1) a pixel-wise L2 loss over the foreground mask $\mathcal{S}$ and the rendered silhouette $\mathcal{M}$, and (2) an edge distance loss, based on the distance of each silhouette boundary pixel $x\in\text{Edge}(\mathcal{M}(M, \mathbf{k}))$ to the nearest foreground mask boundary pixel $\hat{x}\in\text{Edge}(\mathcal{S})$. 

\mheading{SDS loss on normal images} Inspired by \fd~\cite{chen2023fantasia3d}, our approach integrates normal renderings with the SDS loss~\cite{poole2022dreamfusion}.
It enables \mn to effectively capture intricate geometric details without rendering the color image.
Given the surface normals $\mathbf{n}=\mathcal{N}(M, \mathbf{k})$, $\mathcal{L}_\mathrm{SDS}^\mathrm{norm}$ is defined as:
\begin{multline}           
\mathcal{L}_\mathrm{SDS}^\mathrm{norm}= \nabla_{\psi_\mathrm{g}}\mathcal{L}_\mathrm{SDS}^\mathrm{norm}(\mathbf{n}, \mathbf{c}^{P_\mathrm{norm}}) \\
= \mathbb{E}_{\mathrm{t, \epsilon}}\left[ w_t \left(\hat{\epsilon}_{\phi'}(\mathbf{z}_t^\mathbf{n};\mathbf{c}^{P_\mathrm{norm}},t)-\epsilon\right)\frac{\partial \mathbf{n}}{\partial \psi_\mathrm{g}}\frac{\partial \mathbf{z}^\mathbf{n}}{\mathbf{n}}\right], 
\end{multline}
where $\mathbf{c}^{P_\mathrm{norm}}$ is the text condition with an augmented prompt $P_\mathrm{norm}$. 
We construct $P_\mathrm{norm}$ from $P$ by adding an extra description ``a detailed sculpture of'' to better reflect the intrinsic characteristics of normal maps.

\begin{figure}[t]
\centering
\includegraphics[width=\linewidth]{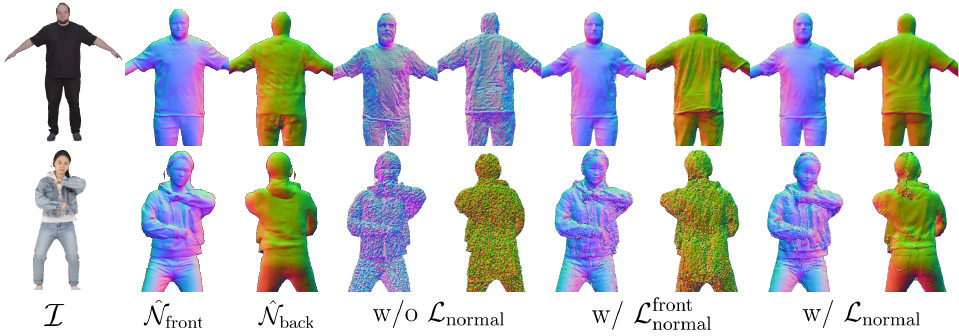}  
\caption{\textbf{The effects of normal regularization.} $\mathcal{L}_\mathrm{norm}$ regularizes the surface with predicted normal images $\hat{\mathcal{N}}_\mathrm{front}, \hat{\mathcal{N}}_\mathrm{back}$.}
\label{fig:ablation_goemtry}
\end{figure}

\mheading{Geometric regularization}
We found that relying solely on silhouette and SDS losses may lead to the generation of noisy surfaces, which is particularly evident for subjects wearing complex clothing.
To address this, we leverage normal estimations as an additional constraint to regularize the reconstructed surface (see~\cref{fig:ablation_goemtry}):
\begin{multline}
    \scriptsize
    \mathcal{L}_\mathrm{norm}(\hat{\mathcal{N}}_\mathbf{k}, \mathbf{n}) 
    = \lambda_\mathrm{MSE}^\mathrm{{norm}} \norm{\hat{\mathcal{N}}_\mathbf{k}- \textbf{n}}_2^2 + \mathrm{LPIPS}(\hat{\mathcal{N}}_\mathbf{k}, \textbf{n})) ,
    \label{loss-normal}
\end{multline}
where $\hat{\mathcal{N}}_\mathbf{k}$ are the front and back normal maps \textit{estimated} using ICON~\cite{xiu2022icon} indexed by the view $\mathbf{k}$ ($\mathbf{k} \in \{\mathrm{front}, \mathrm{back}\}$). $\mathbf{n}$ are the corresponding \textit{differentiably rendered} normal images of the 3D shape $\Psi_\mathrm{g}$.
%
We use a combination of LPIPS and MSE loss to enhance the similarity between $\hat{\mathcal{N}}_\mathbf{k}$ and $\mathbf{n}$. 
Furthermore, we utilize a regularization loss based on Laplacian smoothing~\cite{ando2006learning}, represented as $\mathcal{L}_\mathrm{lap}$.

\mheading{Mesh extraction}
We use Marching Tetrahedra~\cite{doi1991efficient} to extract the mesh from the tetrahedral grid.
Like ECON~\cite{xiu2022econ}, we register \smplx to this mesh which allows us to transfer skinning weights for reposing (see~\cref{fig:application_animation}).
In addition, we replace the hands with \smplx ones which effectively mitigates any potential artifacts introduced during reposing which is needed in the subsequent texture generation stage.
%


\subsubsection{Texture Stage}
\label{sec:texture-stage}
\begin{figure}[t]
\centering
\includegraphics[width=\linewidth]{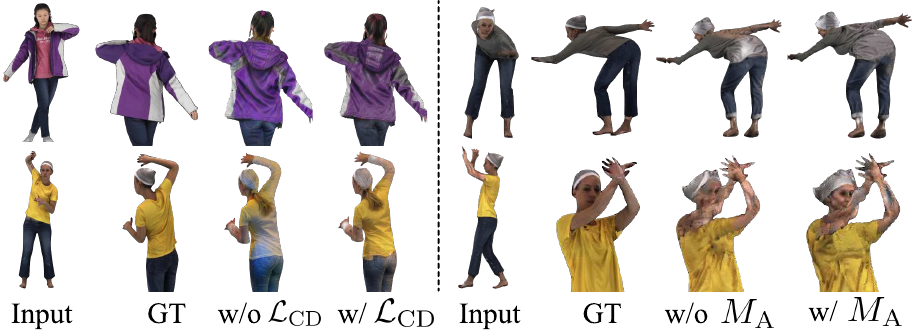}  
\vspace{-1.0 em}
\caption{\textbf{The effects of color consistency loss $\mathcal{L}_\mathrm{CD}$ and multi-pose training ($M_\mathrm{A}$) for texture optimization.}
$\mathcal{L}_\mathrm{CD}$ corrects the over-saturated back-side color generated by SDS, while $M_\mathrm{A}$ improves the texture quality under self-occlusion or extreme poses.
}
\label{fig:ablation_color}
\vspace{-1.0 em} 
\end{figure}

Given the triangular mesh from the geometry stage, we optimize the full texture.
To recover the consistent details and color, even for self-occluded regions, we render both the input pose ($M_\mathrm{in}$) and the A-pose ($M_\mathrm{A}$) during optimization.
%
The textures of $M_\mathrm{in}$ and $M_\mathrm{A}$ are modeled by $\Psi_\mathrm{color}$ in the 3D space of $M_\mathrm{A}$.
We optimize the texture from scratch with $\psi_\mathrm{c}$ randomly initialized.
In~\cref{fig:ablation_color}, we show the effect of this multi-pose training. 
We utilize an occlusion-aware reconstruction loss $\mathcal{L}_\mathrm{recon}$ on the input view of $M_\mathrm{in}$, an SDS loss $\mathcal{L}_\mathrm{SDS}^\mathrm{color}$ with text guidance on rendered color images of both $M_\mathrm{in}$ and $M_\mathrm{A}$,  and a color consistency regularization $\mathcal{L}_\mathrm{CD}$, with respective weights $\lambda$ to balance the individual losses:
\begin{align}
\begin{split}
\label{overall_l_tex}
    \mathcal{L}_\mathrm{texture} &=  \lambda_\mathrm{recon}\mathcal{L}_\mathrm{recon} + \lambda_\mathrm{SDS}\mathcal{L}_\mathrm{SDS}^\mathrm{color} +\lambda_\mathrm{CD}\mathcal{L}_\mathrm{CD} , 
\end{split}
\end{align}
Note that $\mathcal{L}_\mathrm{CD}$ is only utilized after the full-body texture convergence (5000 iters), in an additional optimization phase of 2000 iterations for enforcing color consistency.

\mheading{Occlusion-aware reconstruction loss}
To enforce pixel-alignment, we apply an input view reconstruction loss to minimize the difference between input image $\mathcal{I}$ and the albedo-rendered image $\mathcal{I'}(M, \psi_\mathrm{c}, \mathbf{k_\mathcal{I}})$. Additionally, we have observed that applying $\mathcal{L}_\mathrm{recon}$ to self-occluded areas may lead to incorrect texture due to geometry misalignment. Therefore, an occlusion-aware mask $m_\mathrm{occ}$ is introduced to selectively exclude the $\mathcal{L}_\mathrm{recon}$ in occluded regions.
\begin{align}
\begin{split}
    \mathcal{L}_\mathrm{recon} &= m_\mathrm{occ}(\lambda_\mathrm{MSE} \norm{\mathcal{I} - \mathcal{I'}(M, \psi_\mathrm{c}, \mathbf{k_\mathcal{I}})}_2^2 \\ &+ \mathrm{LPIPS}(\mathcal{I}, \mathcal{I'}(M, \psi_\mathrm{c}, \mathbf{k_\mathcal{I}}))) ,
    \label{recon-loss}
\end{split}
\end{align}
where $\mathbf{k_\mathcal{I}}$ denotes the input view camera, and $\lambda_\mathrm{MSE}$ is a weight to balance the two loss terms. 

\mheading{SDS loss on color images}
To recover the full-body texture, including unseen regions, we update $\psi_\mathrm{c}$ via SDS loss $\mathcal{L}_\mathrm{SDS}^\mathrm{color}$ with text guidance.
This loss is calculated based on random-view color renderings $\mathbf{x}=\mathcal{I'}(\psi_\mathrm{g}, \psi_\mathrm{c}, \mathbf{k})$, and \db $\mathcal{D}'$ parameterized by $\phi'$ and guided by text prompt $P$.
\begin{multline}
    \mathcal{L}_\mathrm{SDS}^\mathrm{color}  =
    \nabla_{\psi_\mathrm{c}}\mathcal{L}_\mathrm{SDS}^\mathrm{color}(\mathbf{x}, \mathbf{c}^P) \\
    = \mathbb{E}_{\mathrm{t, \epsilon}}\left[ w_t \left(\hat{\epsilon}_{\phi'}(\mathbf{z}_t^\mathbf{x};\mathbf{c}^P,t)-\epsilon\right)\frac{\partial \mathbf{x}}{\partial \psi_\mathrm{c}}\frac{\partial \mathbf{z}^\mathbf{x}}{\mathbf{x}}\right] , 
    \label{sds-color-loss}
\end{multline} 
where $\mathbf{k}$ is the camera pose, $\mathbf{c}^P$ is the text embedding of $P$.
%

\mheading{Chamfer-based color consistency loss}
As mentioned in \df~\cite{poole2022dreamfusion}, the SDS loss may result in over-saturated colors which will cause a noticeable color disparity between visible and invisible regions.
To mitigate this issue, we incorporate a color consistency loss to ensure that the rendered novel views align closely with the color distributions observed in the input view.
We quantify the disparity between the color distributions using a chamfer Distance (CD) by treating the pixels from both views as point clouds within the RGB color space:
\begin{multline}
    \mathcal{L}_\mathrm{CD}
    = \sum_{x \in \mathbf{F}_\mathbf{x}} \operatorname*{min}_{y \in \mathbf{F}_\mathcal{I}} ||x-y||^2_2 + \sum_{y \in \mathbf{F}_\mathcal{I}} \operatorname*{min}_{x \in \mathbf{F}_\mathbf{x}} ||x-y||^2_2,
    \label{eq:chamfer_color}
\end{multline}
where $\mathbf{F}_\mathbf{x}$ and $\mathbf{F}_\mathcal{I}$ respectively represent the foreground pixels of the novel-view albedo rendering $\mathbf{x}$, and the input view $\mathcal{I}$. The improvement using $\mathcal{L}_\mathrm{CD}$ is shown in~\cref{fig:ablation_color}.
\subsubsection{Camera sampling during optimization}
\label{sec:camera_sampling}
To optimize the 3D shape and texture using multi-view renderings, cameras are randomly sampled in a way that ensures comprehensive coverage of the entire body by adjusting various parameters.
%
%
%
%
%
To mitigate the occurrence of mirrored appearance artifacts (\ie, Janus-head), we incorporate view-aware prompts (``\specific{front/side/back/overhead view}'') \wrt the viewing angle in the diffusion-based generation process, whose effectiveness has been demonstrated in \db~\cite{poole2022dreamfusion}.
In order to improve facial details, we also sample cameras positioned around the face, together with the additional prompt  ``\specific{face of}''. More details about the camera sampling strategy are in~\cref{sec:camera-sampler} of~\supmat

\section{Experiments} \label{sec:experiments}

\begin{table*}[t]
\centering
\resizebox{\linewidth}{!}{
\begin{tabular}{ccccccccccccc}
\multicolumn{1}{c|}{Method} &
  \multicolumn{6}{c|}{3D Metrics} &
  \multicolumn{6}{c}{2D Image Quality Metrics} \\
\multicolumn{1}{c|}{} &
  \multicolumn{3}{c|}{CAPE} &
  \multicolumn{3}{c|}{\thtwo} &
  \multicolumn{3}{c|}{CAPE} &
  \multicolumn{3}{c}{\thtwo} \\
\multicolumn{1}{c|}{} &
  Chamfer $\downarrow$ &
  P2S $\downarrow$ &
  \multicolumn{1}{c|}{Normal $\downarrow$} &
  Chamfer $\downarrow$ &
  P2S $\downarrow$ &
  \multicolumn{1}{c|}{Normal $\downarrow$} &
  PSNR$\uparrow$ &
  SSIM$\uparrow$ &
  \multicolumn{1}{c|}{LPIPS$\downarrow$} &
  PSNR$\uparrow$ &
  SSIM$\uparrow$ &
  LPIPS$\downarrow$ \\ \shline
\multicolumn{13}{c}{w/o SMPL-X body prior} \\ \hline
\multicolumn{1}{c|}{\pifu~\cite{saito2019pifu}} &
 1.9683  &
  1.6236 &
  \multicolumn{1}{c|}{0.0623} &
  1.9305 &
  1.8031 &
  \multicolumn{1}{c|}{0.0802} & 
  27.0994 &
  0.9362 &
  \multicolumn{1}{c|}{0.0987} &
  23.5068 &
  0.9296 &
  0.1083 \\
\multicolumn{1}{c|}{\pifuhd~\cite{saito2020pifuhd}} &
  3.2018 &
  2.9930 &
  \multicolumn{1}{c|}{0.0758} &
 2.4613  &
  2.3605 &
  \multicolumn{1}{c|}{0.0924} &
  - &
  - &
  \multicolumn{1}{c|}{-} &
- &
- &
-  \\ \hline
\multicolumn{13}{c}{w/ \smplx body prior} \\ \hline
\multicolumn{1}{c|}{\pamir~\cite{zheng2021pamir}} &
  1.3756 &
  1.1852 &
  \multicolumn{1}{c|}{0.0526} &
  1.2979 &
  \textbf{1.2188} &
  \multicolumn{1}{c|}{0.0676} & 
  27.7279 &
  0.9456 &
  \multicolumn{1}{c|}{0.0904} &
  22.5466 &
  0.9266 &
 0.1082 \\
\multicolumn{1}{c|}{\icon~\cite{xiu2022icon}} & 
  0.8689 &
  0.8397 &
  \multicolumn{1}{c|}{0.0360} & 
  \textbf{1.1382} &
  1.2285 & 
  \multicolumn{1}{c|}{0.0623} &
  - &
  - &
  \multicolumn{1}{c|}{-} &
  - &
  - & -
  \\
\multicolumn{1}{c|}{\econ~\cite{xiu2022econ}} & 
  0.9186 &
  0.9227 &
  \multicolumn{1}{c|}{0.0330} & 
  1.2585 &
  1.4184 & 
  \multicolumn{1}{c|}{\textbf{0.0612}} &
  - &
  - &
  \multicolumn{1}{c|}{-} &
  - &
  - & -
  \\ \hline


   
\multicolumn{1}{c|}{\mn} & \textbf{0.7416}
   & \textbf{0.6962}
   &
  \multicolumn{1}{c|}{\textbf{0.0306}} &
  1.2364 &
  1.2715 &

  \multicolumn{1}{c|}{0.0642} &
  \textbf{28.3601} &
  \textbf{0.9490} &
  \multicolumn{1}{c|}{\textbf{0.0639}} &
   \textbf{25.2107} &
   \textbf{0.9363} &
   \textbf{0.0835} \\ 
  
\end{tabular}}
\caption{\textbf{Quantitative evaluation against SOTAs.}
\mn surpasses SOTA baselines in terms of both 3D metrics and 2D image quality metrics. This demonstrates its superior performance in accurately reconstructing clothed human geometry with intricate details, as well as producing high-quality textures with consistent appearance.} 
\vspace{-1.0 em}
\label{table:geo-metrics}
\end{table*}

We compare \mn with state-of-the-art image-based 3D clothed human reconstruction methods, including body-agnostic methods, such as \pifu~\cite{saito2019pifu}, \pifuhd~\cite{saito2020pifuhd} and \phorhum~\cite{alldieck2022photorealistic}, as well as methods that utilize SMPL-(X) body prior, such as \pamir~\cite{zheng2021pamir}, \icon~\cite{xiu2022icon} and \econ~\cite{xiu2022econ}. 
%
%
For a fair comparison, all methods (\ie, \pifu, \pamir, \icon, \econ) utilize the same normal estimator from \icon.
%
%
%
Official \pifu, \pamir and \phorhum are used to evaluate the quality of texture.
For \econ, we use $\text{ECON}_\text{EX}$, due to its superior performance on both ``OOD poses'' and ``OOD outfits'' cases, as reported in the original paper~\cite{xiu2022econ}. 
Note that \phorhum uses a different camera model which is not compatible with our testing data, thus, we use \phorhum only for qualitative comparisons.
More implementation details about network structure and optimization setting can be found at~\cref{sec:implementation} of \supmat

\subsection{Models and Datasets}
\label{sec:experimentsdatasetes}

\qheading{Off-the-shelf models}
\mn relies on multiple off-the-shelf pre-trained models and does not need any additional training data.
Specifically, we use officially released stable-diffusion-v1.5\footnote{\url{runwayml/stable-diffusion-v1-5}} as T2I diffusion model, which is trained on \laion, the VQA model BLIP~\cite{li2022blip} pre-trained on 129M images from multiple datasets~\cite{lin2014coco, krishna2017visual,ordonez2011im2text,ng2020understanding,changpinyo2021cc12m, schuhmann2022laionb} and fine-tuned on VQA2.0~\cite{balanced_vqa_v2}, SegFormer\footnote{\url{matei-dorian/segformer-b5-finetuned-human-parsing}}~\cite{xie2021segformer} pretrained from \cite{cordts2015cityscapes,zhou2017scene,caesar2018coco,deng2009imagenet} and fine-tuned on ATR\cite{ATR}, PIXIE~\cite{feng2021pixie} trained on human images from multiple datasets~\cite{Choutas2020_expose, lin2014coco, Freihand2019, Xiang_2019_CVPR, Parkhi15VGGFace}, and the normal predictor of \icon~\cite{xiu2022icon} trained on AGORA~\cite{patel2020agora}.

\mheading{Datasets for evaluation}
Based on the high-fidelity 3D textured scans from \cape~\cite{CAPE:CVPR:20} and \thtwo~\cite{tao2021function4d}, we perform quantitative evaluations.
We follow ICON~\cite{xiu2022icon} to analyze the robustness of reconstructions under both simple and complex poses (150 scans from \cape).
%
An additional 150 \thtwo scans are included, which comprises 100 subjects that were manually selected to represent a diverse range of clothing styles (\eg, open jackets, long coats, garments with intricate patterns, \etc), and 50 randomly sampled subjects. 
The images are rendered at a resolution of $512\times512$.
For qualitative comparison, we selected the SHHQ dataset~\cite{fu2022styleganhuman} due to its wide range of textures, outfits, and gestures. From this dataset, we randomly sampled 90 images with official mask annotations.

\subsection{Quantitative Comparison}
\label{sec:quantitative-comparison}

We quantitatively evaluate the reconstruction quality of geometry and appearance, using the \textbf{Chamfer} (bi-directional point-to-surface) and \textbf{P2S} (1-directional point-to-surface) distance, to measure the difference between the reconstructed and ground-truth meshes. 
Additionally, we report the L2 \textbf{Normal} error between normal images rendered from both meshes, to measure the consistency and fineness of local surface details, by rotating the camera by $\{0^{\circ}, 90^{\circ}, 180^{\circ}, 270^{\circ}\}$ \wrt to the input view.
To evaluate the quality of the texture, we report 2D image quality metrics, on the multi-view colored images rendered in the same way as the normal images, including \textbf{PSNR} (Peak Signal-to-Noise Ratio), \textbf{SSIM} (Structural Similarity) and \textbf{LPIPS} (learned perceptual image path similarity). 
%
%

%
As shown in~\cref{table:geo-metrics}, \mn demonstrates superior performance across all 2D metrics and 3D metrics on \cape.
This reveals that \mn can accurately reconstruct both geometry and texture, even for subjects with challenging poses (\cape) or loose clothing (\thtwo). 
However, on \thtwo, it achieves comparable reconstruction accuracy to prior-based methods. This can be attributed to the fact that the hallucinated back-side may differ from the ground truth while still appears realistic. A perceptual study~\cref{table:user-study} was conducted for additional clarification.
%
%
%
%
%
See \cref{sec:more-results} of \supmat for more results on these datasets.
%


 \subsection{Perceptual Evaluation}
 \label{sec:qualitative-comparison}
 
\begin{figure*}[ht!]
\centering
\vspace{-1.0 em}
\includegraphics[trim=000mm 000mm 000mm 000mm, clip=True, width=\linewidth]{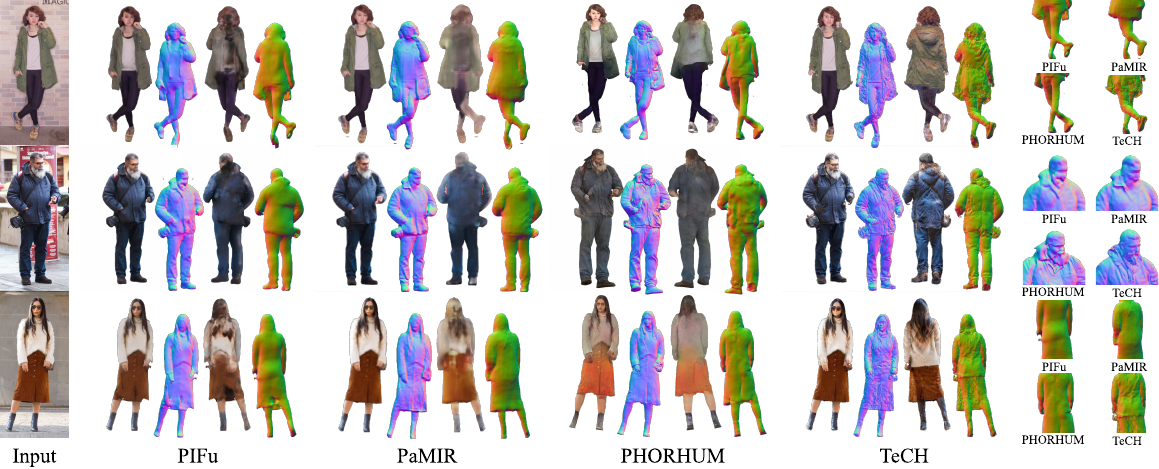}  
\caption{\textbf{Qualitative comparison  on SHHQ images.} \mn generalizes well on in-the-wild images with diverse clothing styles and textures. It successfully recovers the overall structure of the clothed body with text guidance, and generates realistic full-body texture which is consistent with the colored pattern and the material of the clothes. \faSearch~\textbf{Zoom in} to see the geometric details.
}
\label{fig:qualitative-wild}
\end{figure*}

To assess the generalization of \mn on in-the-wild images and evaluate the perceptual quality of our results, we conducted a perceptual study using 90 randomly sampled images from the SHHQ dataset~\cite{fu2022styleganhuman}. 
Participants were shown videos showcasing rotating 3D humans reconstructed by \mn, as well as the baselines (\pamir~\cite{zheng2021pamir}, \pifu~\cite{saito2019pifu}, \icon~\cite{xiu2022icon}, \econ~\cite{xiu2022econ} and \phorhum~\cite{alldieck2022photorealistic}). They were asked to choose the more realistic and consistent result based on the input image.
We gathered a total of 3,150 pairwise comparisons from 63 participants, uniformly covering 90 SHHQ subjects.
The results in \cref{table:user-study} show that \mn is preferred, both, in terms of geometry and texture.
As illustrated in \cref{fig:qualitative-wild}, unlike other methods that tend to reconstruct overly smooth surfaces and blurry textures, \mn shows remarkable generalizability when applied to in-the-wild images featuring diverse clothing styles and gestures.
It produces more realistic clothing, haircut, and facial details, even for unseen back-side views.

\begin{table}[t]
\centering
\scriptsize
\setlength{\tabcolsep}{2.5pt}
\resizebox{1.0\linewidth}{!}{
\begin{tabular}{c|ccc|cc}
Preference (\%, $\uparrow$)           &    \pifu & \pamir & \phorhum  & \icon & \econ\\ \shline
Geometry &   88.6 & 87.0  & 81.7 & 97.94 & 90.48 \\
Colored Rendering  & 95.1 & 93.7  & 93.0 &- &-  \\
\end{tabular}
}
\caption{\textbf{Perceptual study}. The percentages of user preference to \mn compared to other baselines are reported. Most participants preferred \mn in both geometry and colored rendering (texture).}
\label{table:user-study}
\end{table}
%

\subsection{More Qualitative Results}
\label{sec:more-results}
In addition to \cref{fig:qualitative-wild}, we show more qualitative comparisons between \mn and other baselines (\pifu~\cite{saito2019pifu}, \pifuhd~\cite{saito2020pifuhd}, \pamir~\cite{zheng2021pamir}, \phorhum~\cite{alldieck2022photorealistic}, \icon~\cite{xiu2022icon}, \econ~\cite{xiu2022econ}) on CAPE, \thtwo, and SHHQ~\cite{fu2022styleganhuman} images (\cref{fig:qualitative-dataset-cape,fig:qualitative-dataset,fig:qualitative-wild-more} of \supmat), by visualizing multi-view surface normals, color renderings, and zoomed-in details.
For subjects in CAPE and \thtwo, \mn precisely recover the human shape and generate high-quality details of garments and facial features, regardless of hard poses, complex texture, loose clothing, or self-occlusion.
%
%
Also, \cref{fig:qualitative-wild-more} demonstrates the strong generalizability of \mn on in-the-wild images, more rotating 3D humans are provided in \video.

\subsection{Ablation Studies}
\label{sec:ablation}

To assess the effectiveness of key designs in \mn, we perform ablation studies on a 10\% subset of the test set, consisting of 15 subjects from \thtwo and 15 from \cape. The detailed analysis on these results is as follows:

\begin{table}[htbp]
\centering
\scriptsize
\setlength{\tabcolsep}{2.5pt}
\resizebox{\linewidth}{!}{
\begin{tabular}{c|cccccc|ccc|ccc}
& \multicolumn{6}{c|}{Experiment settings} & \multicolumn{3}{c|}{3D Metrics} & \multicolumn{3}{c}{2D Image Quality Metrics}        \\ \shline
& VQA &
  DreamBooth &
  $\mathcal{L}_\mathrm{norm}$ &
  $\mathcal{L}_\mathrm{CD}$ &
  $M_\mathrm{A}$ &
  multi-stage &
  Chamfer $\downarrow$ &
  P2S $\downarrow$ &
  Normal $\downarrow$ &
  PSNR $\uparrow$ &
  SSIM $\uparrow$ &
  LPIPS $\downarrow$ \\ \hline
  \rowcolor[HTML]{EFEFEF}
Ours & \cmark &
  \cmark &
  \cmark &
  \cmark &
  \cmark &
  \cmark &
\bone 0.9794 & \btwo 0.9779 & \btwo 0.0466 &  \btwo 26.7565 &  \bone 0.9428 &  \bone 0.0741 \\ \hline
& \cmark     &  \xmark    & \cmark    & \cmark    & \cmark    & \cmark
& 0.9959         & 1.0192   & \bone 0.0454   & 26.2078                & 0.9405 & 0.0813 \\
A. &  \xmark   & \cmark    & \cmark    & \cmark    & \cmark    & \cmark    &
1.0032   & 1.0218   & 0.0470   & \bone 26.9602               & \bone 0.9428 & 0.0785 \\
& \cmark     & \cmark    & \cmark    & \cmark    & \cmark    &   \xmark   
& \btwo 0.9957         & 0.9963        & 0.0468        &  26.0465               & 0.9395 & 0.0775 \\ \hline
B. & \cmark &
  \cmark & \xmark
   & \xmark & \xmark & \xmark &
  1.0882 & \bone 0.9203 & 0.0870 & - &   - &   - \\ \hline
C. & \cmark     & \cmark    & \cmark    & \cmark    &   \xmark   & \cmark
& -         & -        & -        & 26.6500                & \btwo 0.9427 & \btwo 0.0746 \\
& \cmark     & \cmark    & \cmark    &    \xmark  & \cmark    & \cmark    
& -         & -        & -        & 26.6506                & 0.9425 & 0.0786 \\ 
\end{tabular}}
\caption{\textbf{Ablation study.}
We quantitatively evaluate the effectiveness of each component. Top two results are colored as \colorbox{bestcolor}{first}~\colorbox{secondbestcolor}{second}. All the factors are grouped \wrt their influence: A. geometry+texture, B. geometry only, C. texture only. 
}
\label{table:ablation-metrics}
\end{table}

\mheading{Text guidance} 
%
\Tab{table:ablation-metrics}-A shows that either the ``VQA-only'' or ``\db-only'' guidance exhibit a decrease in performance w.r.t. reconstruction accuracy (Chamfer, P2S) and texture quality (LPIPS).  
%
\Cref{fig:ablation_semantic} shows that VQA prompts help to recover the overall structure of clothing, while \db enhances the fine details of the texture pattern. Combining both text guidance sources yields the best results. A detailed analysis of individual descriptive texts (\eg, garments, hairstyles, \etc) is in~\cref{fig:ablation_semantic_more}


    

\mheading{Geometric regularization}
%
%
As shown in \cref{fig:ablation_goemtry}, using only $\mathcal{L}_\mathrm{SDS}^\mathrm{norm}$ to optimize the geometry will produce noisy artifacts, particularity noticeable in loose clothes. The significant increase in ``Normal'' error shown in~\cref{table:ablation-metrics}-B echos this. This issue can be mitigated by incorporating $\mathcal{\mathcal{L}_\mathrm{norm}}$ at the beginning of the optimization. 

\mheading{Consistent texture recovery}
The results presented in \cref{fig:ablation_color} demonstrate that $\mathcal{L}_\mathrm{CD}$ notably enhances color consistency between the frontal and back sides, and "multi-pose" training ($M_\mathrm{A}$) improves texture quality when dealing with self-occlusion scenarios. This improvement is further supported by \cref{table:ablation-metrics}-C, across all 2D image quality metrics.

\mheading{Multi-stage optimization}
%
As shown in~\cref{table:ablation-metrics}-A, compared to the decoupled two-stage optimization (Ours), the joint optimization results in a performance drop across both 3D and 2D metrics. This may be attributed to the entanglement of the gradients from the geometry and texture branches during optimization.
Notably, in the separate texture stage, a colored image is rendered from the extracted mesh, saving 20\% of the run time compared to joint optimization, which involves rendering from the \dmtet mesh.
\section{Applications}
\label{sec:application}




\subsection{Avatar animation}

\begin{figure}[h]
\centering
\scriptsize
\includegraphics[width=\linewidth]{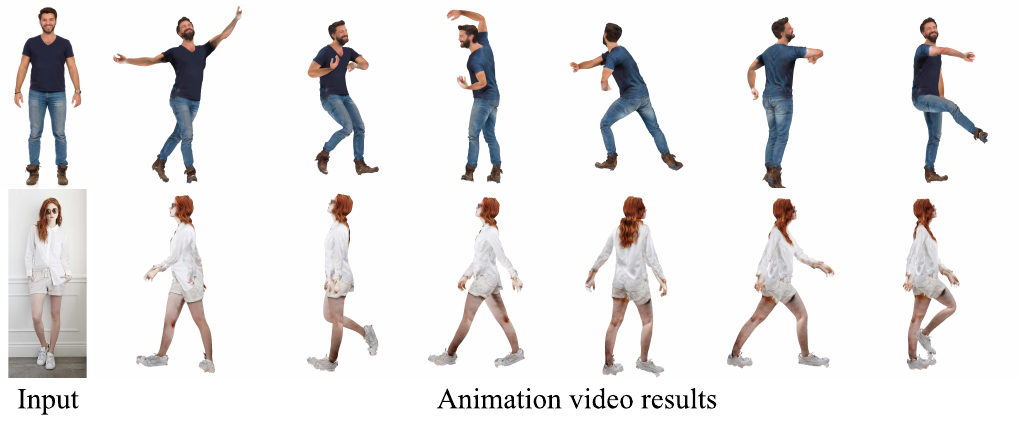}  
\scriptsize
\vspace{-2.0 em}
\caption{\textbf{Animate \mn with \smplx motions.} 
}
\vspace{-1.5 em}
\label{fig:application_animation}
\end{figure}

Following the geometry optimization phase, \mn aligns the clothed body mesh with the \smplx model, enabling us to animate the reconstructed avatar with \smplx motions~\cite{AMASS:ICCV:2019}, as shown in \cref{fig:application_animation} and \video. 

\subsection{Avatar editing}

The text-guided texture generation feature also allows us to edit the texture of the generated avatars.
Here, we show stylization results with different painting styles, like ``\specific{pop art, pixel art, van gogh}''.
The resulting texture not only features the desired styles but also preserves the inherent appearance traits of the original character.

\begin{figure}[h]
\centering
\includegraphics[width=\linewidth]{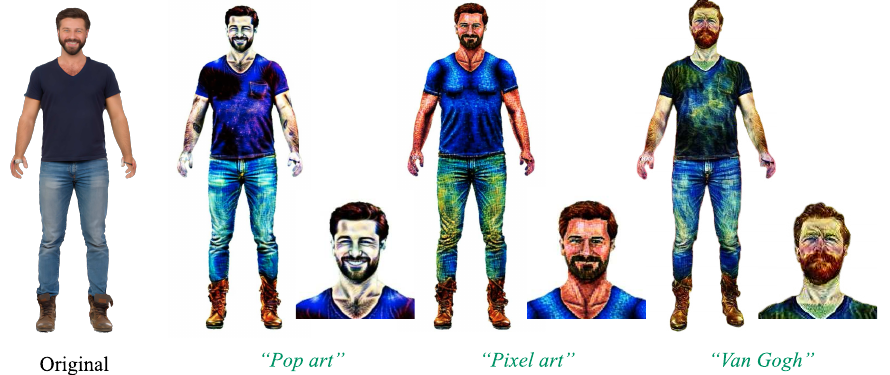}  
\vspace{-2.0 em}
\caption{\textbf{Text-guided stylization.}}
\label{fig:applications}
\vspace{-1.0 em} 
\end{figure} 

\section{Discussion}

\begin{figure}[h]
\centering
\includegraphics[width=\linewidth]{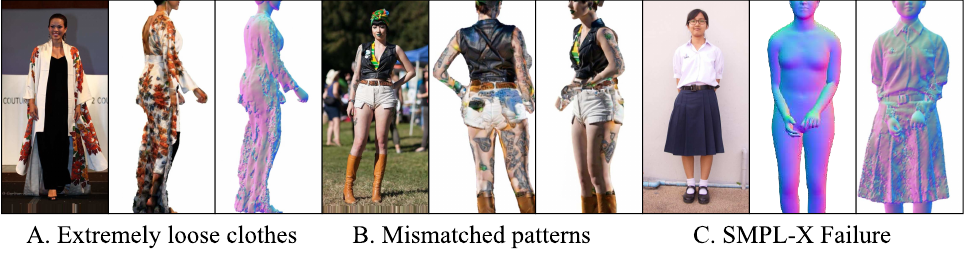}  
\vspace{-2.0 em}
\caption{The proposed method might exhibit noisy surfaces for extremely loose clothing, or mismatched patterns. If PIXIE~\cite{feng2021pixie} is predicting a wrong initial pose, the error propagates to \mn.
}
\label{fig:failure_cases}
\vspace{-1.0 em} 
\end{figure}

\mheading{Limitations}
Despite achieving impressive results on diverse datasets, some failures cases still exist, see~\cref{fig:failure_cases}:
\textbf{A.} \mn occasionally fails for extremely loose clothing, this may relate to the constraint from \smplx-based initialization.
\textbf{B.} mismatched pattern may occur as tattoo.
\textbf{C.} \mn relies on robust \smplx pose estimation, which is still an unsolved problem, especially for challenging poses.

\sheading{Efficiency} For each subject, training \db takes 20 min, \dmtet \smplx initialization takes 20 min, geometry stage (coarse-50 min, fine-50 min), mesh post-processing takes 10 min (remeshing, \smplx registration, hand replacement), texture stage takes 140 min, 270 min in total.
Thus, our per-subject optimization process remains time-consuming, requiring approximately 4.5 hours per subject on a V100 GPU. Addressing these limitations is crucial to facilitate broader applications.

\sheading{Future work}
Leveraging controllable T2I models \cite{ju2023humansd,mou2023t2i,zhang2023adding,oft2023} may help to improve the controllability and stability of generation process. Also, how to compositionally generate the separate components, such as haircut~\cite{sklyarova2023neural}, accessories~\cite{gao2022dart}, and decoupled outfits~\cite{Feng2022scarf}, is still an unsolved problem. We leave these for future research.

\sheading{Broader impact}
\mn has many potential applications~\cref{sec:application}. However, as the technique advances, it has the potential to facilitate deep-fake avatars and raise IP concerns. Regulations should be established to address these issues alongside its benefits in the entertainment industry.

\section{Conclusion} 
\label{sec:conclusion}

We have proposed \mn to reconstruct a lifelike 3D clothed human from a single image, with detailed full-body geometry and high-quality, consistent texture.
The core insight is that we can leverage descriptive text prompts and personalized Text-to-Image diffusion models to optimize the 3D avatar including parts that are not visible in the input.
Extensive experiments validate the superiority of \mn over existing methods in terms of geometry and rendering quality.
We believe that this paradigm of using image and textual descriptions for 3D body reconstruction is a stepping stone also for reconstruction tasks beyond human bodies.


\smallskip
\qheading{\footnotesize{Acknowledgments}} 
{\footnotesize Haven Feng contributes the core idea of ``chamfer distance in RGB space''~\cref{eq:chamfer_color}. We thank Vanessa Sklyarova for proofreading, Haofan Wang, Huaxia Li, and Xu Tang for their technical support, and Weiyang Liu's and Michael J. Black's feedback. Yuliang Xiu is funded by the European Union’s Horizon $2020$ research and innovation programme under the Marie Skłodowska-Curie grant agreement No.$860768$ (\href{https://www.clipe-itn.eu}{CLIPE}). Hongwei Yi is supported by the German Federal Ministry of Education and Research (BMBF): Tubingen AI Center, FKZ: 01IS18039B. Yangyi Huang is supported by the National Nature Science Foundation of China (Grant Nos: 62273302, 62036009, 61936006). Jiaxiang Tang is supported by National Natural Science Foundation of China (Grant Nos: 61632003, 61375022, 61403005).}

\begin{figure*}[htbp]
\centering
\includegraphics[width=\linewidth]{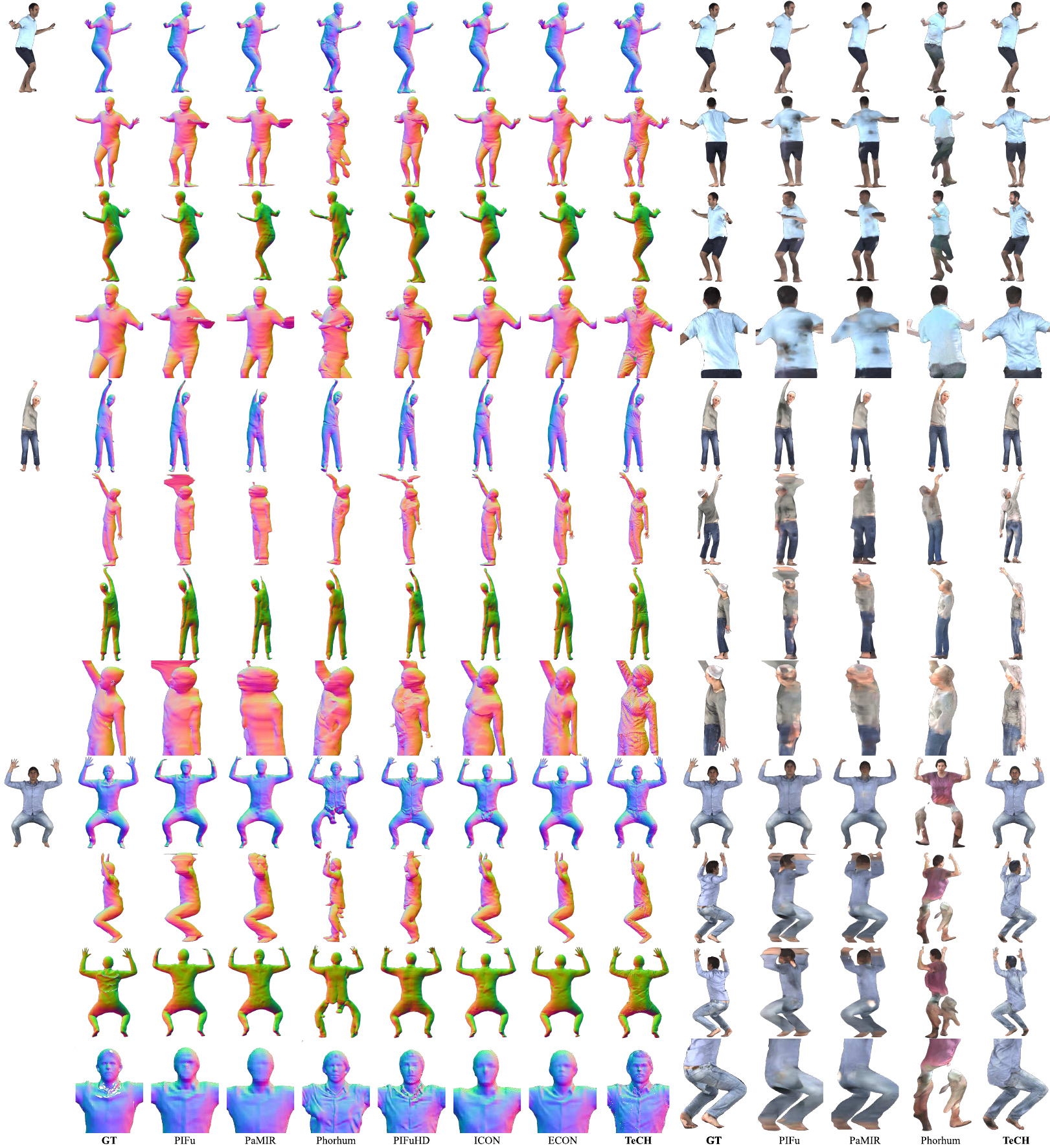}  
\caption{\textbf{Qualitative comparison on \cape}. \mn performs better on subjects with challenging poses.} 
\label{fig:qualitative-dataset-cape}
\vspace{5.0 em} 
\end{figure*}

\begin{figure*}[htbp]
\centering
\includegraphics[width=\linewidth]{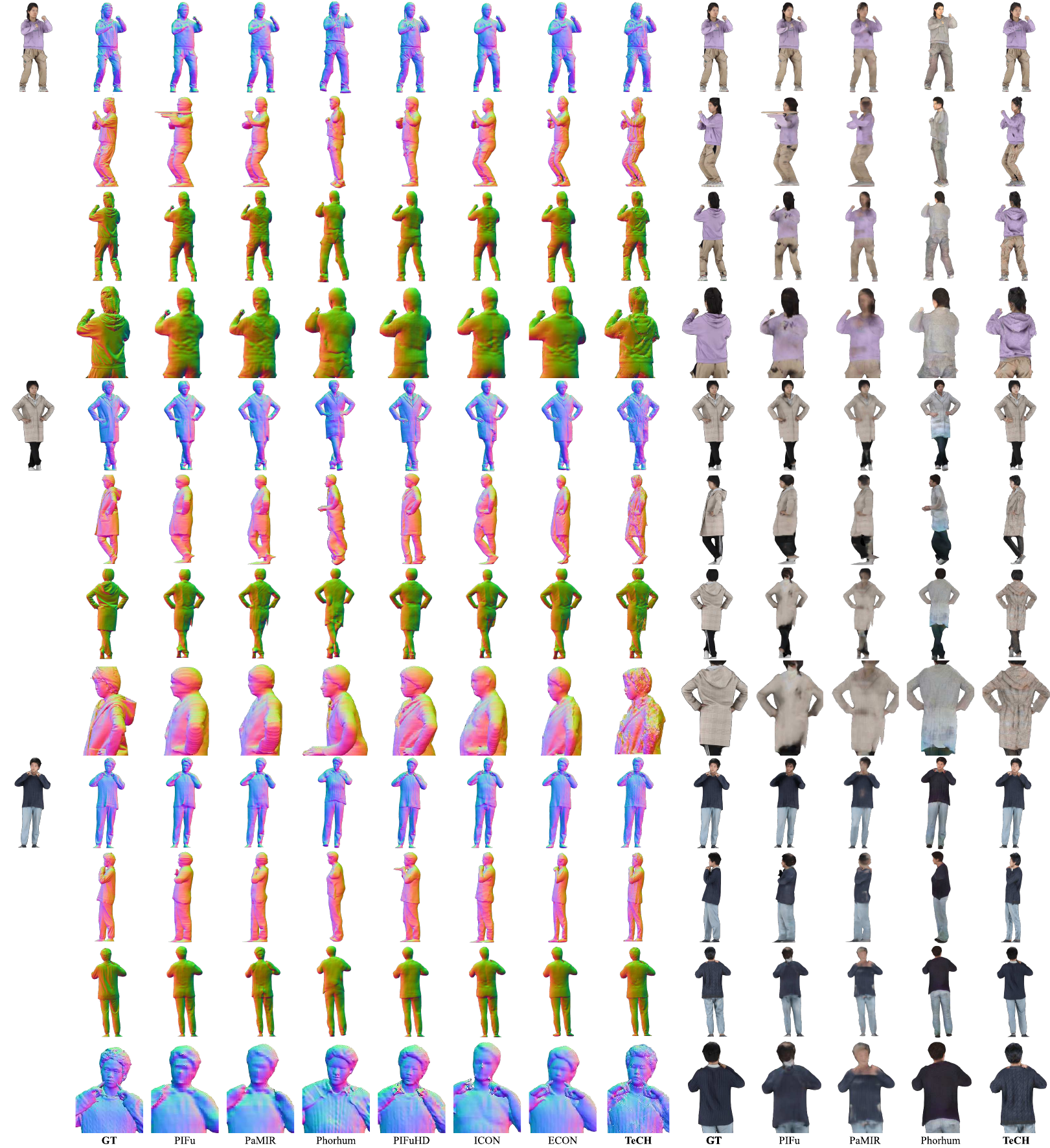}  
\caption{\textbf{Qualitative comparison on \thtwo}. \mn performs better regardless of hard pose, complex texture, or loose clothing.} 
\label{fig:qualitative-dataset}
\vspace{5.0 em} 
\end{figure*}

\begin{figure*}[ht!]
\centering
\includegraphics[trim=000mm 000mm 000mm 000mm, clip=True, width=\linewidth]{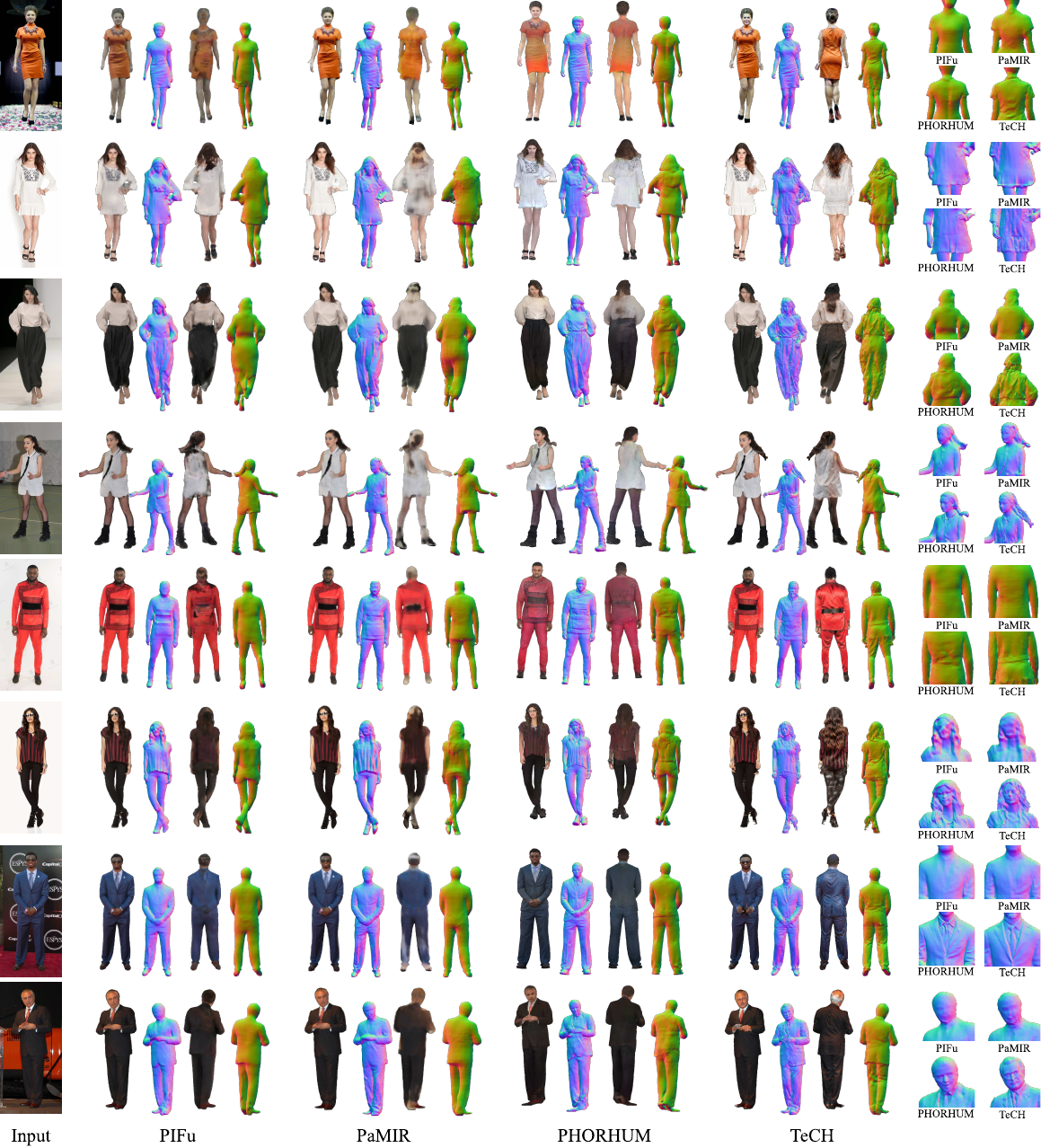}  
\caption{\textbf{Qualitative comparison  on SHHQ images.} \mn generalizes well on in-the-wild images with diverse clothing styles and textures. It successfully recovers the overall structure of the clothed body with text guidance, and generates realistic full-body texture which is consistent with the colored pattern and the material of the clothes. \faSearch~\textbf{Zoom in} to see the geometric details.}
\label{fig:qualitative-wild-more}
\vspace{1.0 em} 
\end{figure*}

\clearpage
\begin{appendices}
\label{appendices}
We provide an additional introduction to the preliminaries (\cref{sec:preliminaries}) of \mn.
%
We list the VQA questions $P_\text{VQA}$ (\cref{sec:vqa-questions}).
%
Additional implementation details to construct the outer shell around \smplx (\cref{sec:outer-shell}), as well as details on the camera sampling strategy (\cref{sec:camera-sampler}) are given.
Implementation details of network structure and optimization setting (\cref{sec:implementation}).
Based on the benchmark datasets (\cape, \thtwo) and in-the-wild photos used in the perceptual studies, we present more qualitative results (\cref{fig:qualitative-dataset,fig:qualitative-dataset-cape,fig:qualitative-wild-more}). 
%

\section{Preliminaries} \label{sec:preliminaries}

\mheading{DreamBooth}
Pretrained text-to-Image diffusion models~\cite{rombach2022high, saharia2022photorealistic, rameshZeroShotTexttoImageGeneration2021} lack the ability to mimic the appearance of subjects in a given reference set and synthesize novel renditions of them in different contexts.
To enable subject-driven image generation, DreamBooth~\cite{ruiz2022dreambooth} \textit{personalizes} the pretrained diffusion model through few-shot tuning.
Specifically, for a pre-trained image diffusion model $\hat{\mathbf{x}}_\phi$, the model takes an initial noise $\epsilon\sim\mathcal{N}(0,1)$, and a text embedding $\mathbf{c}=\Gamma(P)$, generated by the text encoder $\Gamma$ and a text prompt $P$, to produce an image $\mathbf{x}_{\mathrm{gen}}=\hat{\mathbf{x}}_\phi(\epsilon, \mathbf{c})$.
DreamBooth uses 3$\sim$5 images of the same subject to fine-tune the diffusion model using MSE denoising losses:
\begin{multline}
\mathbb{E}_{\mathbf{x},\mathbf{c},\mathbf{\epsilon},\mathbf{\epsilon'},t}  = \bigl[w_t\left\Vert\hat{\mathbf{x}}_\phi(\alpha_t\mathbf{x}_\mathrm{gt}+\sigma_t\mathbf{\epsilon}, \mathbf{c}) - \mathbf{x}_\mathrm{gt}\right\Vert_2^2 \\ + \lambda w_{t'} \left\Vert \hat{\mathbf{x}}_\phi(\alpha_{t'}\mathbf{x}_\mathrm{prior}+\sigma_{t'}\mathbf{\epsilon}', \mathbf{c}_\mathrm{prior})-\mathbf{x}_\mathrm{prior}\right\Vert_2^2\bigr]
\end{multline}
Where $\mathbf{x}_\mathrm{gt}$ represents ground-truth images, and $\mathbf{c}$ is the embedding of a text prompt with a rare token as the unique identifier, and $\alpha_t$, $\sigma_t$, $w_t$ controls the noise schedule and sample quality of the diffusion process at time $t\sim\mathcal{U}([0,1])$.
The second term is the prior-preservation loss weighted by $\lambda$, which is supervised by self-generated images $\mathbf{x}_\mathrm{prior}$ conditioned with the class-specific embedding $\mathbf{c}_\mathrm{prior}=\Gamma(\text{``\specific{a man/woman}''})$.
This loss mitigates the phenomenon of language drift, where the model collapses into a single mode by associating the class name with a particular instance, thus augmenting the output diversity.

\mheading{Score Distillation Sampling (SDS)}
\df~\cite{poole2022dreamfusion} introduces Score Distillation Sampling (SDS) loss, to perform Text-to-3D synthesis by using pretrained 2D Text-to-Image diffusion model $\phi$.
Instead of sampling in pixel space, SDS optimizes over the 3D volume, which is parameterized with $\theta$, with the differential renderer $g$, so the generated image $\mathbf{x}=g(\theta)$ closely resembles a sample from the frozen diffusion model.
Here is the gradient of $\mathcal{L}_\mathrm{SDS}$:
\begin{multline}
\nabla_\theta\mathcal{L}_\mathrm{SDS}(\phi, \mathbf{x}=g(\theta)) \\ = \mathbb{E}_{\mathrm{t, \epsilon}}\left[ w_t \left(\hat{\epsilon}_\phi(\mathbf{z}_t^\mathbf{x};\mathbf{c},t)-\epsilon\right)\frac{\partial \mathbf{x}}{\partial \theta}\frac{\partial \mathbf{z}^\mathbf{x}}{\partial \mathbf{x}}\right]
\end{multline}
where $\hat{\epsilon}_\phi(\mathbf{z}_t^\mathbf{x}; \mathbf{c}, t)$ denotes the noise prediction of the diffusion model with condition $\mathbf{c}$ and latent $\mathbf{z}_t^\mathbf{x}$ of the generated image $\mathbf{x}$.
Such SDS-guided optimization is performed with random camera poses to improve the multi-view consistency. In contrast to \df, the 3D shape here is parameterized with an improved \dmtet instead of NeRF. 

\mheading{Deep Marching Tetrahedra (\dmtet)}
\label{dmtet}
\dmtet~\cite{shen2021deep, gao2020learning} is a hybrid 3D representation designed for high-resolution 3D shape synthesis and reconstruction.
It incorporates the advantages of both explicit and implicit representations, by learning Signed Distance Field (SDF) values on the vertices of a deformable tetrahedral grid.
For a given \dmtet, represented as $(V_T, T)$, where $V_T$ are the vertices in the tetrahedral grid $T$, comprising $K$ tetrahedrons $T_k \in T$, with $k \in \{1,\dots,K\}$.
Each tetrahedron is defined by four vertices $\{v_{k}^1,v_{k}^2,v_{k}^3,v_{k}^4\}$.
The objective of the model is firstly to estimate the SDF value $s(v_i)$ for each vertex, then to iteratively refine the surface and subdivide the tetrahedral grid by predicting the position offsets $\Delta v_i$ and SDF residual values $\Delta s(v_i)$.
A triangular mesh can be extracted through Marching Tetrahedra~\cite{doi1991efficient}.
As noted by Magic3D~\cite{lin2022magic3d}, \dmtet offers two advantages over NeRF, \textbf{fast-optimization} and \textbf{high-resolution}.
It achieves this by efficiently rasterizing a triangular mesh into high-resolution image patches using a differentiable renderer~\cite{Laine2020diffrast}, enabling interaction with pre-trained high-resolution latent diffusion models, such as eDiff-I~\cite{balaji2022eDiff-I}, and \sd~\cite{rombach2022high}.
\section{VQA Questions $Q$}
\label{sec:vqa-questions}

To construct the descriptive prompt $P_\mathrm{VQA}$, we designed a series of questions to parse clothed human attributes.
First, we use \textcolor{orange}{BLIP}~\cite{li2022blip} and a series of general questions $Q_\mathrm{general}$ to parse genders, facial appearance, hair colors, hairstyles, facial hairs, and body poses.
Secondly, we use \textcolor{blue}{SegFormer}~\cite{xie2021segformer} to parse human garments, consisting of 10 categories \specific{\{hat, sunglasses, upper-clothes, skirt, pants, dress, belt, shoes, bag, scarf\}}, denoted as $\textcolor{blue}{G}$, and use another group of questions $Q_\mathrm{garments}$ to parse the attribute of each garment \textcolor{blue}{$g\in G$}.
All the questions are listed in \cref{tab:supmat_vqa_questions}.

\begin{table}[htbp]
\begin{tabular}{ll}
\shline
Groups                       & Quetions                                       \\ \shline
\multirow{8}{*}{$Q_\mathrm{general}$}  & Is this person a man or a woman?               \\
                             & What is this person wearing?                   \\
                             & What is the hair color of this person?         \\
                             & What is the hairstyle of this person?         \\
                             & Describe the facial appearance of this person. \\
                             & Does this person have facial hair?             \\
                             & How is the facial hair of this person?         \\
                             & Describe the pose of this person.              \\ \hline
\multirow{4}{*}{$Q_\mathrm{garments}$} & Is this person wearing \textcolor{blue}{$g$}?                      \\
                             & What \textcolor{blue}{$g$} is the person wearing? $\rightarrow$ \textcolor{orange}{$d$}                 \\
                             & What is the color of the \textcolor{orange}{$d$} + \textcolor{blue}{$g$}?                    \\
                             & What is the style of the \textcolor{orange}{$d$}  + \textcolor{blue}{$g$}?                  
\end{tabular}
\caption{\textbf{Predefined questions for parsing clothed human attributes.} \textcolor{blue}{$g$} is the segmentation category of a part of the garments, and \textcolor{orange}{$d$} is the recognized garment category from the answer to the second question in $Q_\mathrm{garments}$.}
\label{tab:supmat_vqa_questions}
\end{table}

Empirically, we found that the BLIP~\cite{li2022blip} VQA model tends to use 1 $\sim$ 3 words to answer these questions, so we simply concatenate all the answers and remove repeated words to construct $P_\mathrm{VQA}$.
Note that for the CAPE dataset, we add the dataset-specific description ``\specific{hairnet}'' to the guidance as it is hard to be recognized by BLIP.
\section{Construction of the Outer \smplx Shell}
\label{sec:outer-shell}

To construct a compact tetrahedral grid $(V_\mathrm{shell}, T_\mathrm{shell})$, we calculate a coarse outer shell $M_\mathrm{shell}$ from SMPL-X estimated body mesh $M_\mathrm{body}$.
Specifically, we dilate $M_\mathrm{body}$ with an offset of $\Delta M_\mathrm{body}=0.1$ and simplify the mesh by reducing triangle numbers by $r_\mathrm{decimate}=90\%$ using quadric decimation~\cite{hoppe1999new}.
The we generate the tetrahedral grid $(V_\mathrm{shell}, T_\mathrm{shell})$ of this outer shell by TetGen~\cite{hang2015tetgen} with a maximum volume size of $5\times10^{-8}$.
\section{Camera Sampling}
\label{sec:camera-sampler}
To ensure full coverage of the entire body and the human face, during optimization process, we sample virtual camera poses into two groups: 1) $\mathbf{K}_\mathrm{body}$ cameras with a field of view (FOV) covering the full body or the main body parts, and 2) zoom-in cameras $\mathbf{K}_\mathrm{face}$ focusing the face region.
The ratio $\mathcal{P}_\mathrm{body}$ determines the probability of sampling $\mathbf{k}\in \mathbf{K}_\mathrm{body}$, while the height $h_\mathrm{body}$, radius $r_\mathrm{body}$, elevation angle $\phi_\mathrm{body}$, and azimuth ranges $\theta_\mathrm{body}$ are adjusted relative to the \smplx body scale.
Empirically, we set $\mathcal{P}_\mathrm{body}=0.7$, $h_\mathrm{body}=(-0.4, 0.4)$, $r_\mathrm{body}=(0.7, 1.3)$, $\theta_\mathrm{body}=[-180^\circ, 180^\circ)$, $\phi_\mathrm{body}=\{0^\circ\}$, with the $M_\mathrm{body}$ proportionally scaled to a unit space with xyz coordinates in the range $[-0.5, 0.5]$. 
%
%
%
%
To mitigate the occurrence of mirrored appearance artifacts (\ie, Janus-head), we incorporate view-aware prompts, ``\specific{front/side/back/overhead view}'', \wrt the viewing angle during generation process, whose effectiveness has been demonstrated in \db~\cite{poole2022dreamfusion}.

In order to enhance facial details, we sample additional virtual cameras positioned around the face $\mathbf{k}\in \mathbf{K}_\mathrm{face}$, together with the additional prompt  ``\specific{face of}''. 
With a probability of $\mathcal{P}_\mathrm{face} = 1-\mathcal{P}_\mathrm{body} = 0.3$, the sampling parameters include the view target $c_\mathrm{face}$, radius range $r_\mathrm{face}$, rotation range $\theta_\mathrm{face}$, and azimuth range $\phi_\mathrm{face}$.
Empirically, we set $c_\mathrm{face}$ to the 3D position of SMPL-X head keypoint, $r_\mathrm{face}=[0.3, 0.4]$, $\theta_\mathrm{face}=[-90^\circ, 90^\circ]$ and $\phi_\mathrm{face}=\{0^\circ\}$.
%
%

\section{Implementation Details}
\label{sec:implementation}
\subsection{Network Structure}

We use two networks $\Psi_\mathrm{g}$ and $\Psi_\mathrm{c}$ to predict the SDF for geometry modeling and to predict the RGB value for albedo texture modeling, respectively. 
For $\Psi_\mathrm{g}$, we use a 2-layer MLP network with a hidden dimension of 32 and a hash positional encoding with a maximum resolution of 1028 and 16 resolution levels.
During the forward process, we use coordinates of $V_\mathrm{shell}$ in the normalized unit space, the vertices of the tetrahedral grid as the input of $\Psi_\mathrm{g}$ to query SDF value for each vertex.

For $\Psi_\mathrm{c}$, we use a similar network with 1-layer MLP and a hash positional encoding with a maximum resolution of 2048.
We model the albedo texture in the canonical A-pose 3D space. Specifically, for the post-processed result mesh $M_\mathrm{in}=(V_\mathrm{in}, F)$, we register the model with SMPL-X, and repose it with the standard A-pose $M_\mathrm{A}=(V_\mathrm{A}, F)$.
During rendering, if a target pixel is projected onto a triangle $(v_\mathrm{in}^{i}, v_\mathrm{in}^{j}, v_\mathrm{in}^{j}), \text{where} (i,j,k) \in F$ of the $M_\mathrm{in}$. We query the pixel color with its corresponding 3d position in the A-pose space, calculated by interpolation of the triangle $(v_\mathrm{A}^{i}, v_\mathrm{A}^{j}, v_\mathrm{A}^{j})$.
Additionally, we use two 2-layer MLP $\Psi_\mathrm{bg}^\mathrm{g}, \Psi_\mathrm{bg}^\mathrm{c}$ conditioned by camera $\mathbf{k}$ to learn adaptive 3D background colors for both normal map rendering $\mathcal{N}(M, \mathbf{k})$ and color rendering $\mathcal{I'}(M, \psi_\mathrm{c}, \mathbf{k})$.

\subsection{Optimization Details}
%
In both stages of our multi-stage optimization pipeline, we use an Adam optimizer with a base learning rate of $\eta=1\times 10^-3$, and weight decay of $\lambda_\mathrm{WD}=5 \times 10^{-4}$

\mheading{Geometry-stage optimization}
We optimize $\Psi_\mathrm{g}$ in a coarse-to-fine manner, with $t_\mathrm{coarse}=5000$ steps w/o mesh subdivision and $t_\mathrm{fine}=5000$ steps w/ mesh subdivision.
We use a loss weight setting of $\lambda_\mathrm{sil}=1\times 10^4$, $\lambda_\mathrm{SDS}=1$, $\lambda_\mathrm{lap}=1\times 10^4$, and a base loss weight $\lambda_\mathrm{norm}^\mathrm{base}=1\times 10^4$.
For $\lambda_\mathrm{norm}$, to ensure robust convergence of the geometry, we start with a higher value of $\lambda_\mathrm{norm}$ during each stage and gradually decrease it using a two-round cosine annealing, where $\lambda_\mathrm{norm} (t)$ is the weight of $\mathcal{L}_\mathrm{norm}$ at the $t$-th iteration:
\begin{multline}
\lambda_\mathrm{norm} (t) =\\
\begin{cases}
0.5 \lambda_\mathrm{norm}^\mathrm{base} \left(1 + \cos\left(\frac{t}{t_\mathrm{coarse}}\pi\right)\right) & \text{if } t < t_\mathrm{coarse} \\
0.5 \lambda_\mathrm{norm}^\mathrm{base} \left(1 + \cos\left(\frac{t-t_\mathrm{coarse}}{t_\mathrm{fine}}\pi\right)\right) & \text{if } t \geq t_\mathrm{coarse}
\end{cases},
\end{multline}

\mheading{Texture-stage optimization}
We optimize $\Psi_\mathrm{c}$ for $t_\mathrm{texture}=7000$ steps, with $\lambda_\mathrm{recon}=2\times 10^4$ and $\lambda_\mathrm{SDS}=1$.
Besides, we set $\lambda_\mathrm{CD}=0$ at the beginning of the training, and $\lambda_\mathrm{SDS}=1\times 10^6$ at the last $t_\mathrm{CD}=2000$ iterations to enforce color consistency.


\end{appendices}
\clearpage

{
    \small
    \bibliographystyle{ieeenat_fullname}
    \bibliography{main}
}

\end{document}